\documentclass{article}

\usepackage[preprint]{neurips_2026}

\usepackage[utf8]{inputenc} 
\usepackage[T1]{fontenc}    
\usepackage{url}            
\usepackage{booktabs}       
\usepackage{amsfonts}       
\usepackage{nicefrac}       
\usepackage{microtype}      
\usepackage{xcolor}         
\usepackage{amsmath}
\usepackage{floatflt}
\usepackage{bm}
\usepackage{graphicx}
\usepackage{colortbl} 
\usepackage{enumitem} 
\usepackage{wrapfig}
\usepackage{listings}
\usepackage{algorithmic}  
\usepackage{multirow} 
\usepackage{amssymb}
\usepackage{pifont} 
\usepackage{newfloat}

\usepackage{listings}

\newtheorem{proposition}{Proposition}

\usepackage{bbm}
   
\usepackage[most]{tcolorbox}
\usepackage{newfloat}
\usepackage{listings}
\usepackage{caption}
\usepackage{enumitem}
\usepackage[ruled,vlined]{algorithm2e}

\definecolor{morandiblue}{RGB}{145,168,176}
\definecolor{morandired}{RGB}{196,150,148}
\definecolor{mydarkred}{RGB}{170,30,30}

\usepackage{arydshln}   
\usepackage[dvipsnames,table]{xcolor}  
\usepackage[
    colorlinks=true,
    linkcolor=Maroon,    
    citecolor=RoyalBlue,  
    urlcolor=magenta      
]{hyperref}
\title{\texttt{ReSum}: Synergizing LLM Reasoning and Summarization with Reinforcement Learning}

\author{%
  Xucong Wang\thanks{\parbox[t]{0.85\linewidth}{%
    Equal Contribution.  \\ \textsuperscript{\textdagger}Project lead: Yong Wang; Corresponding authors: Yong Wang and Pengkun Wang}}\quad 
  Ziyu Ma$^{*}$\quad 
  Yong Wang$^{\dagger}$\quad 
  Shidong Yang\quad 
  Hailang Huang\quad  \\ 
  \textbf{Renda Li\quad   
  Pengkun Wang$^{\dagger}$\quad 
  Xiangxiang Chu}\\
   AMAP, Alibaba Group  
}

\begin{document}

\maketitle

\begin{abstract}
Reinforcement Learning with Verifiable Rewards (RLVR) is a central technique for improving long-horizon reasoning in Large Language Models (LLMs). However, existing RLVR methods often encourage unnecessarily long reasoning rollouts, which can degrade reasoning coherence and exhaust the available context budget. Existing approaches to long-context organization often depend on external mechanisms to organize rollouts, rather than enabling the model to manage its own reasoning trajectory. To address this limitation, we propose ReSum, a novel RLVR framework that enables LLMs to compress and organize their reasoning trajectories through self-summarization. Our pilot studies show that self-summarization stabilizes generation by lowering token-level entropy, and that introducing a ``summarization'' phrase can substantially mitigate errors propagated from an incorrect rollout prefix. Motivated by these findings, ReSum adopts a summarization-aware adaptive rollout mechanism that contrastively evaluates whether self-summarization benefits the ongoing reasoning process. Specifically, when the model spontaneously triggers self-summarization, ReSum masks the summarization phrase to create a contrastive branch; for non-summarization positions, it instead randomly injects the phrase to create a matched branch. We further design a summarization-aware advantage to enable finer-grained comparison between contrastive rollout trajectories. Extensive experiments show that ReSum improves performance at an average of 4\% while reducing  rollout length by 18.6\%. GitHub Page: \href{https://github.com/xuc865/Resum}{\textbf{GitHub}}.
\end{abstract}

\vspace{-0.2cm}
\section{Introduction}
Recently, Reinforcement Learning with Verifiable Rewards (RLVR)~\cite{schulman2017proximal,guo2025deepseek,yu2025dapo,li2025search} has substantially improved the long-chain reasoning abilities of Large Language Models (LLMs) across complex tasks, including search~\cite{jin2025search,li2025search,wang2025stepsearch}, coding~\cite{mai2025agent,liu2024rl}, and GUI interaction~\cite{shi2025mobilegui,lu2026ui,yuan2025enhancing,luo2025gui}. However, these gains often come with a reward bias toward longer Chain-of-Thought (CoT) sequences~\cite{chen2024not,zhang2025llms,liu2026badthink,wu2025more,srivastava2025llms}, which makes LLMs prone to overthinking. As reasoning trajectories become longer, models may forget or misremember earlier steps~\cite{dong2025longred,liu2024forgetting,chen2024long}, revisit already resolved subproblems, and exhaust their reasoning budget through repeated retries~\cite{ling2025longreason} or self-validation loops.

To address these issues, existing studies have explored a range of long-context organization techniques~\cite{koh2022empirical,dai2025pretraining,ming2024faitheval}, which compress long interaction histories into summaries through cache monitoring~\cite{yang2024pyramidinfer}, autoencoder-based compressors~\cite{ge2023context,dai2025pretraining}, or additional agentic modules~\cite{zhang2025agentic,wu2025resum}. However, these methods typically rely on intricate workflows or auxiliary models to manage long contexts externally, which may introduce additional overhead and raise concerns about the faithfulness~\cite{longpre2021entity,ming2024faitheval,long2025copy} of summaries or outputs produced by external modules. This  motivates a central question: \textit{can LLMs' own self-summarization serve as an intrinsic mechanism for organizing long reasoning trajectories?} To investigate this question, we conduct two pilot studies, as shown in Figure~\ref{fig:pilot}. In Figure~\ref{fig:pilot}(a), inspired by recent studies~\cite{gao2025beyond,dong2025agentic} that highlight the importance of high-entropy tokens in LLM reasoning, we analyze token entropy around summarization phrases in rollout trajectories. We find that positions preceding summarization phrases consistently exhibit high entropy, while the summarization phrase itself is followed by a sharp entropy decrease. In Figure~\ref{fig:pilot}(b), we truncate incorrect rollouts at different positions and regenerate 100 continuations with and without an injected summarization phrase. We find that injecting summarization phrases consistently improves regeneration accuracy, with gains of up to 30\%. Together, these findings reveal a previously overlooked behavior of LLMs:

\begin{figure*}[t] 
    \centering
    \includegraphics[width=1\linewidth]{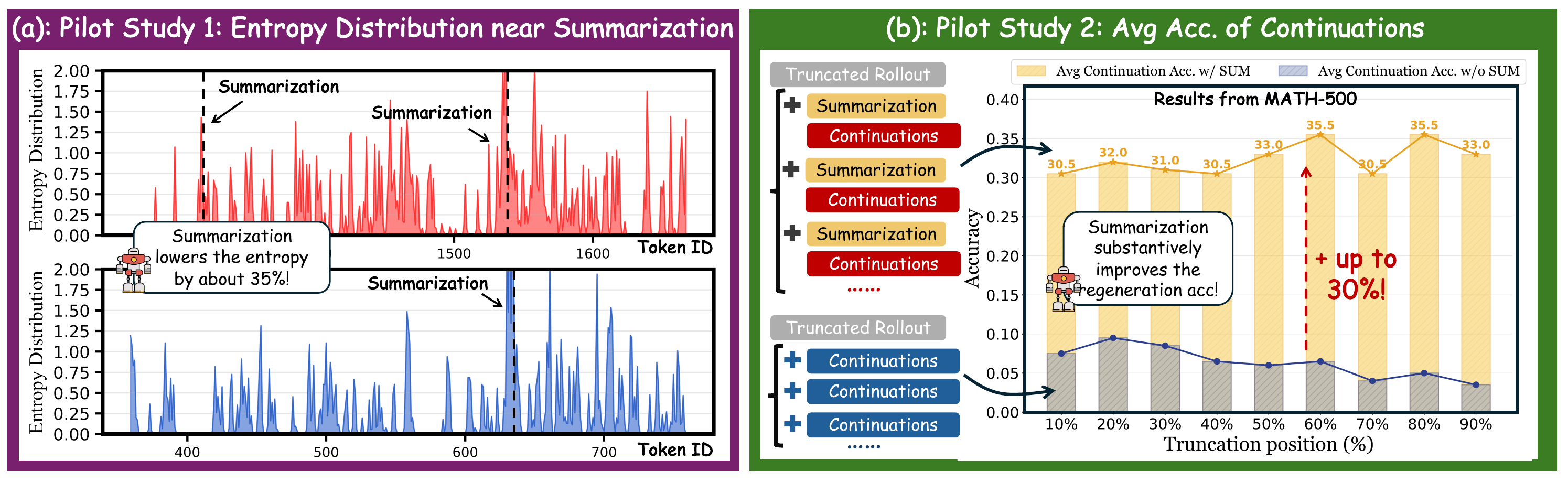} 
    \vspace{-0.6cm}
    \caption{Pilot studies. \textbf{(a):} Analysis of token entropy distributions near the summarization phrases. \textbf{(b):} Average accuracy of continuations of wrong rollouts truncated from different positions.} 
    \vspace{-0.5cm}
    \label{fig:pilot}
\end{figure*}

\begin{tcolorbox}[
  colback=gray!10,     
  colframe=black,    
  boxrule=1pt,       
  arc=2mm,         
  left=4pt,right=4pt,top=4pt,bottom=4pt, 
]
 \textit{Self-summarization serves as an intrinsic control mechanism in long-horizon reasoning: it arises when the model reaches a high-entropy state, and reduces generation uncertainty by consolidating ongoing rollouts and reorganizing the wrong prefixes.}
\end{tcolorbox}

Motivated by these findings, we train LLMs to proactively self-summarize at appropriate points in the reasoning process, thereby compressing accumulated context and recovering from low-quality rollout prefixes. To this end, we propose ReSum, an RL framework that encourages beneficial self-summarization through adaptive branching and fine-grained credit assignment. ReSum builds contrastive rollout branches from two complementary types of points. \textbf{(1) Artifact Points (APs)} are sampled from non-summarization positions, where a summarization phrase is injected to estimate whether summarization would improve the subsequent continuation. \textbf{(2) Natural Points (NPs)} correspond to spontaneous self-summarization behaviors produced by the model, where the summarization phrase is masked to evaluate whether the natural summary contributes to the final outcome. Together, these two branching strategies enable ReSum to learn both when summarization should be introduced and whether an existing summarization behavior should be reinforced.

On the resulting rollout tree, ReSum partitions branches into summarized and non-summarized groups, and computes both intra-group and inter-group advantages for fine-grained credit assignment. Among summarized rollouts, the intra-group advantage favors trajectories that place summaries at more effective points and achieve higher final rewards. In contrast, the inter-group advantage compares summarized branches against non-summarized ones, estimating the task-specific utility of summarization itself. In this way, ReSum achieves fine-grained process supervision for self-summarization behaviors. Without any supervised fine-tuning, ReSum improves accuracy by an average of 4\% while reducing rollout length by 18.6\%. Our contributions are threefold:
\vspace{-0.1cm}
\begin{itemize}[leftmargin=*]
\vspace{-0.1cm}
\item We present pilot studies showing that \textit{self-summarization} stabilizes LLM reasoning by lowering token-level uncertainty and mitigating error propagation from incorrect rollout prefixes.
\vspace{-0.1cm}
\item We propose ReSum, a tree-based RL framework that incentivizes beneficial self-summarization during reasoning. ReSum combines two types of contrastive branching with summarization-aware advantage to provide fine-grained supervision over whether to summarize and when to summarize.
\vspace{-0.1cm}
\item Extensive experiments validate the effectiveness of ReSum, which outperforms existing approaches by an average of 4\% across various benchmarks while substantially reducing rollout length.
\end{itemize}

\section{Related Work}
\paragraph{Reinforcement Learning with Verifiable Rewards.} Beyond prompting and supervised fine-tuning, Reinforcement Learning with Verifiable Rewards (RLVR)~\cite{schulman2017proximal,rafailov2023direct,guo2025deepseek,liu2025understanding} has emerged as a core driver for enhancing the long chain-of-thought reasoning capabilities of LLMs and incentivizing reasoning behaviors such as planning and reflection. Building upon traditional Proximal Policy Optimization (PPO)~\cite{schulman2017proximal}, subsequent research has focused on designing critic-free, lightweight, and efficient policy gradient paradigms, which can be roughly grouped into three lines: \textbf{(1) fine-grained objective designs}~\cite{guo2025deepseek,yu2025dapo,zheng2025group,wang2026textbf}, which incorporate finer advantage or loss formulations~\cite{guo2025deepseek,zhao2025geometric,feng2025group,chu2025gpg}, apply policy-gradient losses at multiple granularities~\cite{yu2025dapo,zheng2025group}, or adopt curriculum scaling~\cite{dai2026harder,li2026adacurl}. For instance, DAPO~\cite{yu2025dapo} extends the gradient loss from the sequence level in GRPO~\cite{guo2025deepseek} to token-level variants, while GPG~\cite{chu2025gpg} returns to the original policy gradient formulation and removes auxiliary clipping or KL-regularization terms. \textbf{(2) process-level rewarding}~\cite{ma2026fipo,feng2025group}, which tackles the credit assignment problem  by distributing sequence-level rewards with a discount factor~\cite{ji2025tree,xie2024monte} or by training Process Reward Models (PRMs)~\cite{shao2024deepseekmath,zhang2025process,wang2025stepsearch}. \textbf{(3) rollout expansion}~\cite{ji2025tree,hou2025treerl,dong2025agentic,dong2025agentic1}, which assigns distinct advantages and gradient losses to rollout segments of varying importance. For example, TreeRL~\cite{hou2025treerl} selects tokens with the highest uncertainty for branching, while Tree-GRPO~\cite{ji2025tree} selects thought-action steps and optimizes them with a two-level advantage.
\vspace{-0.2cm}
\paragraph{In-context Organization for LLMs.} As emerging studies unraveling that \textit{``longer reasoning is not always better''}~\cite{wu2025more,liu2026badthink,srivastava2025llms,chen2024not}, recent work has discovered the in-context organization~\cite{li2025upfront,petrov2025long} (like compression~\cite{berton2025compllm}, resolve~\cite{wang2023plan}, self-correction~\cite{xiong2025self,zhao2025boosting,paul2024refiner} and summarization) mechanisms to refine the models' attention of the long in-contexts~\cite{hooper2025squeezed}. Based on the invoking mode of such techniques, there are \textit{\textbf{(1)} external methods}~\cite{dai2025pretraining,ge2023context,guo2026less}, where the long-contexts are reframed by cache monitors~\cite{ma2026compressing}, text compressors~\cite{tian2026keepkv} or harnessing engineering~\cite{zhang2025agentic,ma2026skillclaw}. \textit{\textbf{(2)} internal methods}~\cite{wang2026textbf,bensal2025reflect,li2025encouraging,li2025reflectevo}, where models are trained or incentivized to selectively revisit their generated content with distillation, supervised fine-tuning or RLVR. For example, Re$^2$~\cite{wang2026textbf} encourages LLMs to perform a "redo" operation with an out-group reward and Test-Time Scaling (TTS)~\cite{muennighoff2025s1}; RLTR~\cite{li2025encouraging} leverages the ``tool-call completeness'' as the reward to refine the planning ability of agents to mitigate the insufficient annotations in real-world applications. \textit{In alignment with these studies, this paper studies how summarization ability can be internalized for complex reasoning.}
\vspace{-0.2cm}
\section{Methodology}
\subsection{Preliminaries} 
\noindent\textbf{Notations.} We adopt the standard RLVR notation for Large Language Models (LLMs). The LLM $\pi_\theta$ serves as the policy model and autoregressively generates rollouts given an input query $q\in\mathcal{D}$, where $\mathcal{D}$ is the training set. For the rollouts $\{o_m\}_{m=1}^{M}$ generated by the current policy $\pi_\theta$, we assign a scalar reward $r_m$ using either LLM-based reward models or verifiable reward functions. In the case of mathematical reasoning, $r_m\in\{0,1\}$ indicates verifiable task reward of mathematical results.

\noindent\textbf{Group Relative Policy Optimization (GRPO).} GRPO~\cite{guo2025deepseek} is a critic-free policy optimization method that computes relative advantages within a group of responses sampled from the same query. It has inspired numerous subsequent studies aimed at refining the normalization scheme~\cite{chu2025gpg}, adjusting levels of gradient optimization~\cite{yu2025dapo}, weighting different samples or rollouts~\cite{dai2026harder}, and related aspects. Specifically, GRPO optimizes the following objective to train the policy model $\pi_\theta$:
\begin{equation}
\begin{split}
        \mathcal{J}_{\rm GRPO}(\theta)=&\mathbb{E}_{q\sim\mathcal{D},\{o_m\}_{m=1}^M\sim\pi_{\rm old}(\cdot;q)}[\frac{1}{M}\sum_{m=1}^M\frac{1}{|o_m|}\sum_{i=1}^{|o_m|}{\rm min}\big(\rho_{m,i}(\theta)\hat{A}_{m,i}, \\
        &{\rm clip}(\rho_{m,i}(\theta),1-\epsilon,1+\epsilon)\hat{A}_{m,i}\big)-\beta\mathbb{D}_{\rm KL}(\pi_\theta(\cdot;q)||\pi_{\rm ref}(\cdot;q))] \\
        \rho_{m,i}(\theta)=&\frac{\pi_\theta(o_{m,i};o_{m,<i},q)}{\pi_{\rm old}(o_{m,i};o_{m,<i},q)}, \ \ \ \hat{A}_{m,i}=\frac{r_m-{\rm mean}(\{r_m\}_{m=1}^M)}{{\rm std}(\{r_m\}_{m=1}^M)}  \\
\end{split}
\end{equation}
Where $\pi_{\rm old}$ and $\pi_{\rm ref}$ represent the old policy and reference policy respectively. $\mathbb{D}_{\rm KL}$ represents the KL-Divergence. $\epsilon$ and $\beta$ are the clipping range and regularization hyper-parameter respectively. $\rho_{m,i}$ represents the importance sampling ratio of rollout $o_m$ at token $o_{m,i}$. 

\subsection{Summarization-based Rollout Expansion}
Beyond binary correctness of the final answer, periodic summarizations help mitigate erroneous outcomes caused by repetitive retries or the forgetting of historical context. While explicitly injecting summarization instructions into the model prompt is a viable option, we cannot guarantee the stability of the model's instruction-following capability over time, especially without cold-start training. We therefore propose ReSum, which leverages RLVR to foster the emergence of spontaneous summarization behaviors from within the model during the rollout process, shown in Figure~\ref{fig:main}. (A detailed comparison between naive prompt refinement and ReSum is provided in Appendix~\ref{appx:e}.)

Inspired by recent work showing that tree-based RLVR, i.e., sampling multiple branches from a single rollout, is equivalent to preference alignment between branches~\cite{zhang2025process,lai2024step}, we devise a tree-based rollout strategy grounded in summarization behavior. By dynamically injecting or masking summarization phrases and incorporating branches into advantage groups, ReSum enables the model to spontaneously learn \textit{where and when to summarize}. Specifically, the tree generation follows:

\noindent\textbf{$\blacktriangleright$ Initialization:} For each query $q$, we first generate $T$ independent initial rollouts $\{\tau_t^{0}\}_{t=1}^T$ from the policy model $\pi_\theta$. Each initial rollout $\tau_t^{0}$ serves as the root of a rollout tree $\mathcal{T}_t$.

\vspace{-0.1cm}
\noindent\textbf{$\blacktriangleright$ Sampling:} In the $j$-th generation loop, where $j=1,\ldots,J-1$, for each rollout tree $\mathcal{T}_t$, we select one point from two types of branching points: Artifact Points (APs) $\mathcal{L}^{ap}_{t,j}$ and Natural Points (NPs) $\mathcal{L}^{np}_{t,j}$. APs are randomly sampled from the initial rollout and used to artificially inject summarization signals. NPs are located at the model's spontaneous summarization identified via keyword matching. Notably, we prioritize the selection of NPs; APs are only used to fill the remaining budget after all of the NPs have been selected.

\vspace{-0.1cm}
\noindent\textbf{$\blacktriangleright$ Branching:} In the $j$-th generation loop, for each rollout tree $\mathcal{T}_t$, we resample branches from either $\mathcal{L}^{ap}_{t,j}$ or $\mathcal{L}^{np}_{t,j}$. \textbf{(1)} When branching at $\mathcal{L}^{ap}_{t,j}$, we obtain the prefix $\mathcal{F}^{ap}_{t,j}$, then \textbf{append} a summarization phrase (i.e., \texttt{[In summary]}) sampled from a collection of summarization-like expressions. Details of this collection are provided in Appendix~\ref{appx:d}. The policy model is then required to regenerate the continuation $\mathcal{O}^{ap}_{t,j}$ accordingly, forming a complete branch $\tau^{ap}_{t,j}=\mathcal{F}^{ap}_{t,j}\circ\mathcal{O}^{ap}_{t,j}$. \textbf{(2)} When branching at $\mathcal{L}^{np}_{t,j}$, we obtain the prefix $\mathcal{F}^{np}_{t,j}$, then \textbf{mask} the identified natural summarization phrases from the prefix. The policy model is then required to regenerate the continuation $\mathcal{O}^{np}_{t,j}$, forming a complete branch $\tau^{np}_{t,j}=\mathcal{F}^{np}_{t,j}\circ\mathcal{O}^{np}_{t,j}$. We merge $\tau^{ap}_{t,j}$ (or $\tau^{np}_{t,j}$) into $\mathcal{T}_t$ as $\mathcal{T}_t\leftarrow\mathcal{T}_t\cup\{\tau^{ap}_{t,j}\}$ (or $\mathcal{T}_t\leftarrow\mathcal{T}_t\cup\{\tau^{np}_{t,j}\}$).

\vspace{-0.1cm}
\noindent\textbf{$\blacktriangleright$ Termination:} The overall generation loop terminates when the total number of rollouts reaches the budget $B$, where $B$ is divisible by $T$. Notably, branches are evenly distributed across trees.

\begin{figure*}[t] 
    \centering
    \includegraphics[width=1\linewidth]{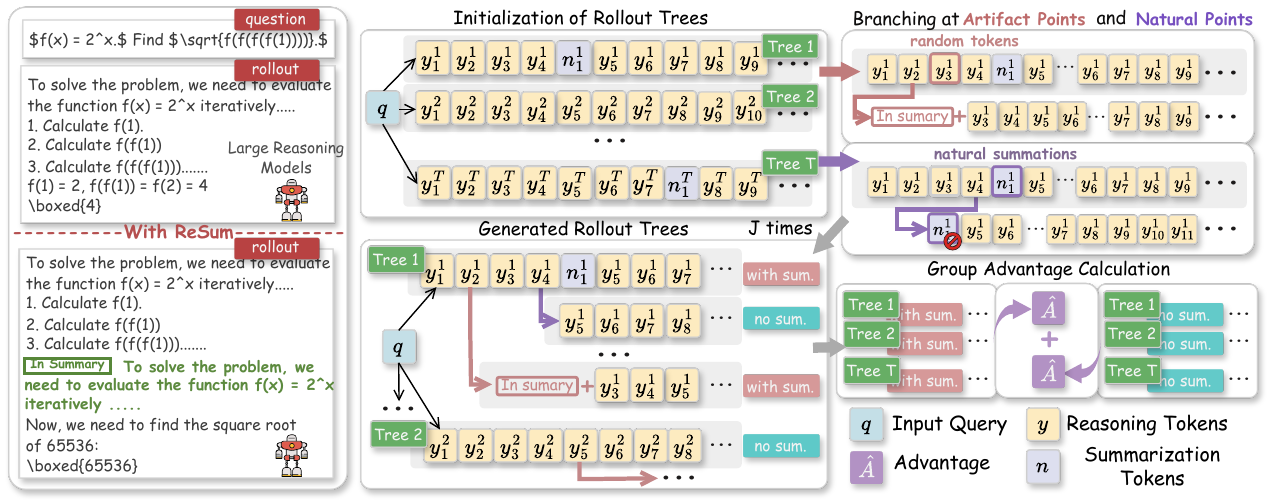} 
    \vspace{-0.5cm}
    \caption{The proposed ReSum. \textbf{(left):} ReSum incentivizes the summarization behavior of the LLM during reasoning. \textbf{(middle \& right)}: the LLM generates initial rollouts for each query, then adopt the rollout branching at both Artifact Points (APs) and Natural Points (NPs). The advantage is calculated within branches with and without the summarization respectively.} 
    \vspace{-0.5cm}
    \label{fig:main}
\end{figure*} 

Different combinations of $T$ and $J$ can yield rollout trees of varying shapes. Specifically, a larger $T$ leads to greater diversity in summarizations across different rollouts, while a larger $J$ produces more branches within each individual rollout. We examine this trade-off further in the ablation studies. Notably, all branching operations are performed exclusively on the initial rollouts.
 
\subsection{Summarization-aware Group-relative Policy Optimization (SGPO)}\label{sec:33}
In tree-based rollouts, the trajectories formed by different branches naturally provide process-level supervision signals. That is to say, summarizations that lead to a higher proportion of correct rollouts naturally receive higher advantage values. Building on this, we design a summarization-aware group advantage mechanism to encourage the model to reinforce high-quality summarization behaviors. Specifically, let $\tau_{t,e}$ denote the $e$-th rollout in $\mathcal{T}_{t}$. We define the reward $R$ as $R(\tau_{t,e})=R_{A}(\tau_{t,e})+R_F(\tau_{t,e})$, where the task reward $R_{A}$ and the format reward $R_{F}$ are defined as follows:
\begin{equation}\label{eq:reward}
 R_{A}(\tau_{t,e})=
\begin{cases}
1, & \!\!\! \text{if } \ {\rm Correct}(\tau_{t,e})\\
0, & \!\!\! \text{otherwise}
\end{cases}\!, 
R_F(\tau_{t,e})= \frac{0.2\!\times\! N_{\rm NP}(\tau_{t,e})+0.05\!\times\! N_{\rm AP}(\tau_{t,e})}{N_{\rm NP}(\tau_{t,e})+N_{\rm AP}(\tau_{t,e})}
\end{equation}
where $N_{\rm NP}(\tau_{t,e})$ and $N_{\rm AP}(\tau_{t,e})$ denote the number of NPs and APs in the rollout. Our reward design favors rollouts with more spontaneous summarizations regardless of their final correctness, assigning an extra reward of 0.2 for each. For artifact summarizations injected at APs, we assign a 0.05 reward as an encouragement for different summarization timings. Finally, for each rollout, these rewards are weighted based on the number of times APs and NPs appear in the rollout, in order to avoid encouraging endless summarizations. We then compute the advantages separately for rollout sets with and without summarization, denoted as $G_{\rm sum}$ and $G_{\rm nosum}$:
\begin{equation}\label{eq:sumnosum}
    \hat{A}^{\rm sum/nosum}(\tau_{t,e})=\frac{R(\tau_{t,e})-{\rm mean}_{\tau\in G_{\rm sum/nosum}}(\{R(\tau)\})}{{\rm std}_{\tau\in G_{\rm sum/nosum}}(\{R(\tau)\})}
\end{equation}
For a rollout that contains summarization, $\hat{A}^{\rm sum}$ encourages the model to spontaneously learn the optimal timing for summarization, by measuring the relative return gaps among multiple rollouts that also exhibit summarization behaviors. Conversely, $\hat{A}^{\rm nosum}$ compares the overall return distribution between rollouts with and without summarization to establish an absolute baseline; this prevents reward hacking and ensures that the overall quality of responses is not significantly compromised. Similarly, for a rollout with no summarization included, this grouping design can also provide fine-grained supervision signals. The overall rollout-level advantage is then computed as $\hat{A}(\tau_{t,e})=\hat{A}^{\rm sum}(\tau_{t,e})+\hat{A}^{\rm nosum}(\tau_{t,e})$ and assigned to all tokens in $\tau_{t,e}$; Accordingly the optimization objective of ReSum is finally given by:
\begin{equation}\label{eq:obj}
\begin{split}
        \mathcal{J}_{\rm ReSum}(\theta)=&\mathbb{E}_{q\sim\mathcal{D},\{\mathcal{T}_t\}_{t=1}^T\sim\pi_{\rm old}(\cdot;q)}
        [\frac{1}{T}\sum_{t=1}^T\frac{1}{|\mathcal{T}_t|}\sum_{e=1}^{|\mathcal{T}_{t}|}\frac{1}{|\tau_{t,e}|}\sum_{i=1}^{|\tau_{t,e}|}{\rm min}\big(\rho_{t,e,i}(\theta)\hat{A}(\tau_{t,e}), \\
        &{\rm clip}(\rho_{t,e,i}(\theta),1-\epsilon,1+\epsilon)\hat{A}(\tau_{t,e})\big)-\beta\mathbb{D}_{\rm KL}(\pi_\theta(\cdot;q)||\pi_{\rm ref}(\cdot;q))]
\end{split}
\end{equation}
Where $\pi_{\rm old}$ and $\pi_{\rm ref}$ represent the old policy and reference policy respectively. $\mathbb{D}_{\rm KL}$ represents the KL-Divergence. $\epsilon$ and $\beta$ are the clipping range and regularization hyper-parameter respectively. $\rho_{t,e,i}(\theta)=\pi_\theta(\tau_{t,e,i};\tau_{t,e,<i},q)/\pi_{\rm old}(\tau_{t,e,i};\tau_{t,e,<i},q)$ represents the importance sampling ratio. 

\subsection{Theoretical Perspective}
\label{sec:theory}

Inspired by the tree-structured analysis in recent studies~\cite{ji2025tree,zhao2026training}, we provide a concise justification for ReSum. The core idea is that summarization acts as a context compression operator, which removes redundant history while preserving the information needed for future reasoning. This makes the reasoning process easier to optimize under a fixed rollout budget.

Specifically, let $H_{<i}$ denote the prefix before the token $i$, and let $\mathcal{S}(H_{<i})$ be the prefix concatenated with the summarization words. Let $p^\star$ denote the target continuation distribution. Assume that the continuation distribution under $\mathcal{S}(H_{<i})$ is close to that under $H_{<i}$, i.e., the summarization is faithful and accurate. We define the compression error as $\Delta_i = \mathbb{D}_{\rm KL}\!\left(p^\star(\cdot \mid q, H_{<i}) \,\|\, p^\star(\cdot \mid q, \mathcal{S}(H_{<i}))\right)$. In ReSum, we branch the rollout into two kinds of paths from the same prefix: one with summarization and one without. Let their expected returns be denoted by $R_{\mathrm{sum}}(i)$ and $R_{\mathrm{nosum}}(i)$, respectively. We define the summary gain as
$G_i = R_{\mathrm{sum}}(i) - R_{\mathrm{nosum}}(i)$. If summarization preserves task-relevant information while reducing redundant context, then the summary branch should achieve a non-negative gain on average.  Furthermore, we introduce the following proposition:

\begin{proposition}[Summarization-aware advantage]
\label{prop:1}
Assume that (i) summarization is $\delta$-faithful, i.e., $\Delta_i \le \delta$, and (ii) the reward is locally stable with respect to small changes in the continuation distribution. Then the summary branch does not degrade the expected return, and strictly improve it when the removed context is mostly redundant.
\end{proposition} 

The detailed derivations of the distributional bound are deferred to Appendix~\ref{appx:a}. This proposition offers a simple explanation for our empirical results: ReSum learns when to summarize, thereby reducing overthinking, improving credit assignment, and yielding more useful training signals under the same rollout budget.
\section{Experiments}
\subsection{Experiment Setup}
\paragraph{Datasets and Backbones.} For training, we employ the MATH~\cite{hendrycks2021measuring} dataset following DGPO~\cite{dai2026harder}, which consists of problems with varying difficulty levels and requires long-chain reasoning. For evaluation, we select six commonly used mathematical benchmarks: AIME24, AIME25, AMC23, Minerva~\cite{lewkowycz2022solving}, MATH500~\cite{hendrycks2021measuring}, and Olympiad~\cite{he2024olympiadbench}. Results on AIME24, AIME25, and AMC23 are averaged over 32 runs, while results on the remaining benchmarks are averaged over 4 runs. We select Qwen2.5-Math-7B~\cite{yang2024qwen2}, Qwen2.5-Math-1.5B~\cite{yang2024qwen2}, Qwen2.5-3B~\cite{yang2412qiu}, and DeepSeek-Math-7B~\cite{shao2024deepseekmath} as our backbone models. We also apply ReSum to the multimodal benchmark GEOQA-8K~\cite{chen2025vinci} to further validate its generality, with Qwen2.5-3B-VL-Instruct~\cite{bai1others} as the backbone.

\vspace{-0.3cm}
\paragraph{Baselines.} We select several competitive baselines for comprehensive comparison, including GRPO~\cite{guo2025deepseek}, Dr. GRPO~\cite{liu2025understanding}, GPG~\cite{chu2025gpg}, DAPO~\cite{yu2025dapo}, GSPO~\cite{zheng2025group}, GRPO-AD~\cite{zhang2025grpo}, and DGPO~\cite{dai2026harder}. To ensure a fair comparison, we make the following adjustments to the baselines: \textbf{(1)} we disable the resampling design in GPG and DAPO; \textbf{(2)} we retain only the advantage-reweighting component from GRPO-Lead~\cite{zhang2025grpo}, which we refer to as GRPO-AD; \textbf{(3)} we disable the Multi-Aspect Reformulation (MQR) in MathForge~\cite{dai2026harder} and treat only its core loss design, termed Difficulty-aware Group Policy Optimization (DGPO), as our baseline. Details of these baselines are provided in Appendix~\ref{appx:c}.

\vspace{-0.3cm}
\paragraph{Implementation Details.} Following DGPO~\cite{dai2026harder}, our codebase is built on Open-R1~\cite{face2025open}. We use 8 NVIDIA H20 GPUs for all experiments. During training, the response temperature is set to 1.0. During evaluation, the response temperature and the top-$p$ value are set to 0.6 and 0.95, respectively. The maximum completion length is set to 4096 for all stages. Further implementation details are provided in Appendix~\ref{appx:d}. The significance test of ReSum against DGPO is reported in Appendix~\ref{appx:e}.

\begin{table}[t]
\caption{Performance comparison between ReSum and others on 6 challenging mathematical reasoning datasets, using Qwen2.5-Math-7B. The best results are in \textbf{bold}. }
\centering
\label{tab:my-table:1}
\renewcommand{\arraystretch}{1.1}   
\setlength\tabcolsep{7pt} 
\resizebox{1\columnwidth}{!}{
\begin{tabular}{l|cccccc|c}
\toprule[1pt]
Method & AIME24 & AIME25 & AMC23 & MATH500 & Minerva & Olympiad & Avg./$\Delta_{\rm GRPO}$ \\ \hline
Base Model & 12.19 & 4.79 & 35.23 & 48.60 & 15.07 & 16.33 & 22.04 \\
GRPO & 20.94 & 8.44 & 58.98 & 72.20 & 27.76 & 37.33 & 37.61 \\
Dr. GRPO & 21.04 & 8.23 & 58.59 & 72.05 & 28.58 & 35.89 & 37.40\textcolor{red}{\tiny -0.21} \\
GPG & 21.98 & 9.06 & 59.61 & 72.05 & 27.21 & 37.67 & 37.93\textcolor[rgb]{0,0.4,0}{\tiny +0.32} \\
DAPO & 21.25 & 8.75 & 58.20 & 72.70 & 29.50 & 37.22 & 37.94\textcolor[rgb]{0,0.4,0}{\tiny +0.33} \\
GSPO & 19.38 & 8.33 & 60.16 & 73.00 & 28.12 & 37.26 & 37.71\textcolor[rgb]{0,0.4,0}{\tiny +0.10} \\
GRPO-AD & 21.56 & 9.48 & 59.06 & 73.25 & 29.14 & 37.07 & 38.26\textcolor[rgb]{0,0.4,0}{\tiny +0.65} \\
DGPO & 23.85 & 10.21 & 61.02 & 74.25 & 31.07 & 38.33 & 39.79\textcolor[rgb]{0,0.4,0}{\tiny +2.18} \\
\rowcolor{gray!10}
\textbf{ReSum (ours)} & \textbf{25.42} &	\textbf{13.33} &	\textbf{62.50} 	&\textbf{76.45} &	\textbf{32.44} &	\textbf{39.67} &	\textbf{41.64}\textcolor[rgb]{0,0.4,0}{\tiny +4.03}  \\ \bottomrule[1pt]
\end{tabular}}\vspace{-0.6cm}
\end{table}

\begin{table}[t]
\caption{Performance comparison between ReSum and others on 6 challenging mathematical reasoning datasets, using various backbones. The best results are in \textbf{bold}. }
\centering
\label{tab:my-table:2} 
\renewcommand{\arraystretch}{1.1}   
\setlength\tabcolsep{5pt} 
\resizebox{1\columnwidth}{!}{
\begin{tabular}{l|ccccccc}
\toprule[1pt]
Method & \multicolumn{1}{l}{AIME24} & \multicolumn{1}{l}{AIME25} & \multicolumn{1}{l}{AMC23} & \multicolumn{1}{l}{MATH500} & \multicolumn{1}{l}{Minerva} & \multicolumn{1}{l}{Olympiad} & \multicolumn{1}{l}{Avg./$\Delta_{\rm GRPO}$} \\ \hline
Qwen2.5-Math-1.5B & 6.87 & 3.65 & 30.94 & 34.95 & 8.55 & 21.93 & 17.82 \\
\,\,+GRPO & 11.35 & 3.96 & 46.48 & 64.85 & 20.13 & 29.59 & 29.39 \\
\,\,+DGPO & 11.25 & 5.73 & 49.84 & 65.45 & 21.14 & 30.85 & 30.71\textcolor[rgb]{0,0.4,0}{\tiny +1.32} \\
\rowcolor{gray!10}
\,\,+\textbf{ReSum (ours)} & \textbf{15.96} & \textbf{6.83 }& \textbf{51.33} & \textbf{67.12} & \textbf{24.58} & \textbf{32.59} & \textbf{33.07}\textcolor[rgb]{0,0.4,0}{\tiny +3.68} \\ \hline
Qwen2.5-3B & 2.81 & 0.73 & 22.66 & 48.65 & 13.69 & 19.37 & 17.99 \\
\,\,+GRPO & 5.31 & 1.56 & 33.28 & 63.35 & 22.89 & 26.41 & 25.47 \\
\,\,+DGPO & 6.98 & 1.56 & 36.56 & 65.80 & 25.28 & 26.96 & 27.19\textcolor[rgb]{0,0.4,0}{\tiny +1.72} \\
\rowcolor{gray!10}
\,\,+\textbf{ReSum (ours)} & \textbf{7.50 } & \textbf{2.29 }& \textbf{39.06} & \textbf{68.40} & \textbf{26.69} & \textbf{28.93} & \textbf{28.81}\textcolor[rgb]{0,0.4,0}{\tiny +3.34} \\ \hline
DeepSeek-Math-7B & 0.42 & 0.10 & 13.28 & 31.05 & 9.56 & 9.00 & 10.57 \\
\,\,+GRPO & 0.63 & 0.10 & 19.14 & 41.45 & 14.71 & 13.44 & 14.91 \\
\,\,+DGPO & \textbf{1.98 }& 0.42 & 21.02 & 41.85 & 18.93 & 15.00 & 16.53\textcolor[rgb]{0,0.4,0}{\tiny +1.62} \\
\rowcolor{gray!10}
\,\,+\textbf{ReSum (ours)} & 1.66 & \textbf{0.96 }&\textbf{22.40} & \textbf{43.10} & \textbf{20.22}  & \textbf{15.56}  &  \textbf{17.32}\textcolor[rgb]{0,0.4,0}{\tiny +2.31} \\ \bottomrule[1pt]
\end{tabular}}\vspace{-0.4cm}
\end{table}

\begin{table}[t]
\caption{Comparison of synergistic results of DGPO / ReSum with other policy optimization methods using Qwen2.5-Math-7B. The best results are in \textbf{bold}. }
\centering
\label{tab:my-table:3} 
\renewcommand{\arraystretch}{1.1}   
\setlength\tabcolsep{7pt} 
\resizebox{1\columnwidth}{!}{
\begin{tabular}{l|cccccc|c}
\toprule[1pt]
Method & AIME24 & AIME25 & AMC23 & MATH500 & Minerva & Olympiad & Avg. / $\Delta$ \\ \hline
GPG & 21.98 & 9.06 & 59.61 & 72.05 & 27.21 & 37.67 & 37.93 \\
\,\,+DGPO & 21.77 & 10.00 & 60.00 & 73.45 & 30.06 & 38.26 & 38.92\textcolor[rgb]{0,0.4,0}{\tiny +0.99} \\
\rowcolor{gray!10}
\,\,+\textbf{ReSum (ours)} & \textbf{27.29} & \textbf{10.83} & \textbf{61.88} & \textbf{74.75} & \textbf{30.76} & \textbf{39.07} & \textbf{40.76}\textcolor[rgb]{0,0.4,0}{\tiny +2.83} \\ \hline
DAPO & 21.25 & 8.75 & 58.20 & 72.70 & 29.50 & 37.22 & 37.94 \\
\,\,+DGPO & 24.48 & 9.79 & 58.75 & 74.90 & 31.99 & \textbf{39.56} & 39.91\textcolor[rgb]{0,0.4,0}{\tiny +1.97} \\
\rowcolor{gray!10}
\,\,+\textbf{ReSum (ours)} & \textbf{25.67} & \textbf{10.67} & \textbf{62.18} & \textbf{75.30} & \textbf{32.11} & 39.34 & \textbf{40.88}\textcolor[rgb]{0,0.4,0}{\tiny +2.94} \\ \hline
GSPO & 19.38 & 8.33 & 60.16 & 73.00 & 28.12 & 37.26 & 37.71 \\
\,\,+DGPO & 23.33 & \textbf{10.00} & 59.14 & 74.15 & 30.88 & 38.41 & 39.32\textcolor[rgb]{0,0.4,0}{\tiny +1.61} \\
\rowcolor{gray!10}
\,\,+\textbf{ReSum (ours)} & \textbf{25.10} &	\textbf{10.00} &	\textbf{61.04} &	\textbf{74.50} &	\textbf{31.38} &	\textbf{38.98} &	\textbf{40.17}\textcolor[rgb]{0,0.4,0}{\tiny +2.46}  \\ \bottomrule[1pt]
\end{tabular}}\vspace{-0.4cm}
\end{table}

\begin{table}[t]
\caption{Performance comparison between ReSum and others on multimodal dataset GEOQA-8K with Qwen2.5-3B-VL-Instruct. The best results are in \textbf{bold}. }
\centering
\label{tab:my-table:4} 
\renewcommand{\arraystretch}{1.1}   
\setlength\tabcolsep{2pt} 
\resizebox{1\columnwidth}{!}{
\begin{tabular}{l|c|ccccccc|c}
\toprule[1pt]
Method & Base Model & GRPO & Dr. GRPO & GPG & DAPO & GSPO & GRPO-AD & DGPO & \textbf{ReSum} \\ \hline
Avg. / $\Delta_{\rm GRPO}$ & 39.79 & 57.43 & 57.96\textcolor[rgb]{0,0.4,0}{\tiny +0.53} & 59.02\textcolor[rgb]{0,0.4,0}{\tiny +1.59} & 59.02\textcolor[rgb]{0,0.4,0}{\tiny +1.59} & 57.16\textcolor[rgb]{0.4,0,0}{\tiny -0.27} & 58.09\textcolor[rgb]{0,0.4,0}{\tiny +0.66} & 59.95\textcolor[rgb]{0,0.4,0}{\tiny +2.52} & \textbf{62.04}\textcolor[rgb]{0,0.4,0}{\tiny +5.61} \\ \bottomrule[1pt]
\end{tabular}}\vspace{-0.4cm}
\end{table}

\begin{table*}[t]
\caption{Experiment results under different rollout budgets $B=T\times J$ and configurations of tree structures with Qwen2.5-3B. The best results are in \textbf{bold}.}
\vspace{-0.1cm}
\centering
\label{tab:my-table:budget} 
\renewcommand{\arraystretch}{1.1}   
\setlength\tabcolsep{6pt} 
\resizebox{1\columnwidth}{!}{
\begin{tabular}{lccccccc}
\toprule[1pt]
\multicolumn{1}{l}{Method} & AIME24 & AIME25 & AMC23 & MATH500 & Minerva & Olympiad & Avg./$\Delta_{\rm GRPO}$ \\ \hline
\multicolumn{1}{l}{Qwen2.5-3B} & {2.81} & {0.73} & {22.66} & {48.65} & {13.69} & {19.37} & {17.99} \\
\multicolumn{1}{l}{GRPO} & {5.31} & {1.56} & {33.28} & {63.35} & {22.89} & {26.41} & {25.47} \\ \hline
\multicolumn{8}{c}{\textit{Rollout Budget $B=T\times J=2$}} \\ \hline 
\multicolumn{1}{l}{ReSum $(T=2,   J=1)$} & 4.67 &	1.56 	&33.06 &	62.73 &	22.81 &	24.96 &	24.97\textcolor[rgb]{0.4,0,0}{\tiny -0.50}   \\ \hline 
\multicolumn{8}{c}{\textit{Rollout Budget $B=T\times J=4$}} \\ \hline 
\multicolumn{1}{l}{ReSum $(T=2,   J=2)$} & 5.57 &	1.86 &	35.97 &	66.90 &	23.81 &	26.41 &	26.75\textcolor[rgb]{0,0.4,0}{\tiny +1.28}   \\
\multicolumn{1}{l}{ReSum $(T=4,   J=1)$} & 6.13 &	2.07 &	37.50& 	65.76 &	25.53 &	27.79 &	27.46\textcolor[rgb]{0,0.4,0}{\tiny +1.99}  \\ \hline
\multicolumn{8}{c}{\textit{Rollout Budget $B=T\times J=16$}} \\ \hline
\multicolumn{1}{l}{ReSum $(T=2,   J=8)$} & 6.67 &	2.10 	&37.10 &	67.97 	&27.05 &	26.97 &	27.98\textcolor[rgb]{0,0.4,0}{\tiny +2.51}  \\
\multicolumn{1}{l}{ReSum $(T=8,   J=2)$} & 6.60 	&\textbf{2.84 }&	38.11 &	68.17 &	26.33 	&\textbf{29.25} &	28.55\textcolor[rgb]{0,0.4,0}{\tiny +3.08}  \\
\rowcolor{gray!20}
\multicolumn{1}{l}{ReSum $(T=4,   J=4)$} & \textbf{7.50 }&	2.29 	&\textbf{39.06} &	\textbf{68.40} &	\textbf{26.69} 	&28.93 &	\textbf{28.81}\textcolor[rgb]{0,0.4,0}{\tiny +3.34}  \\ \bottomrule[1pt]
\end{tabular}}\vspace{-0.3cm}
\end{table*}

\begin{figure*}[t] 
    \centering
    \includegraphics[width=1\linewidth]{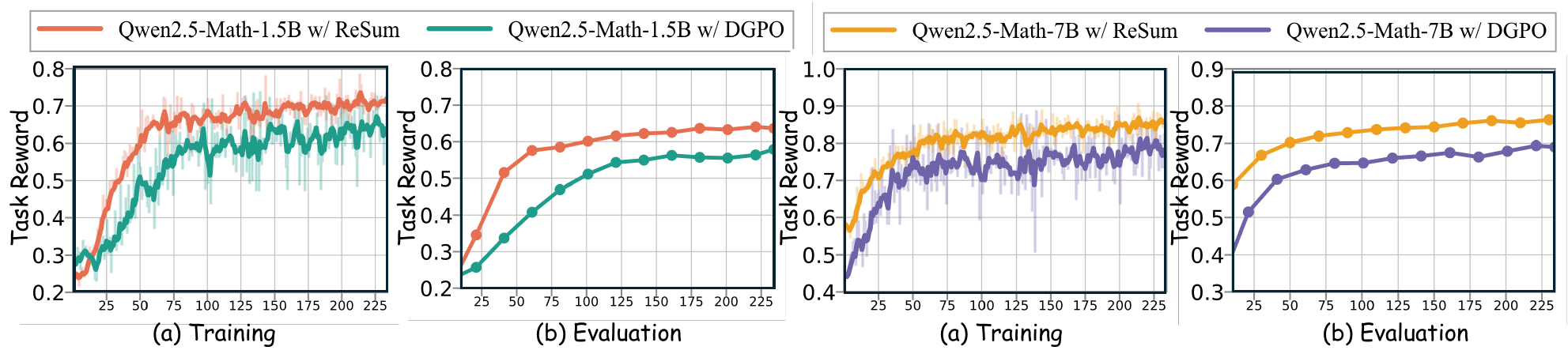} 
    \vspace{-0.5cm}
    \caption{The training and evaluation dynamics of the task reward. We compare ReSum and DGPO with Qwen2.5-Math-1.5B \textbf{(left)} and Qwen2.5-Math-7B \textbf{(right)}.} 
    \vspace{-0.6cm}
    \label{fig:dyna}
\end{figure*}

\subsection{Main Results}
\paragraph{Comparison with State-of-the-Art Methods.}  
Table~\ref{tab:my-table:1} summarizes the accuracy of ReSum and several competitive RLVR baselines on six mathematical benchmarks, all using Qwen2.5-Math-7B as the backbone. ReSum achieves the highest scores on all datasets, reaching an average of 41.64\% and performance gain of 4.03\% over standard GRPO. This trend is particularly evident on the competitive datasets: For example on AIME24, ReSum outperforms DGPO by 1.57\%, and on AIME25 ReSum reaches 13.33\% compared to 10.21\%. We think that while the reweighting schemes from DGPO also improve upon GRPO, it operates primarily on outcome-level signals. In contrast, ReSum explicitly incentivizes the model to condense its own intermediate context, directly mitigating the memory decay and repetitive self-verification that accumulate over long rollouts. The uniform improvements confirm that self-summarization is a broadly effective reasoning strategy.

\vspace{-0.2cm}
\paragraph{Application to Different Backbones.} 
We further examine whether the gains observed with ReSum are transferable across model sizes and families. Table~\ref{tab:my-table:2} reports results with Qwen2.5-Math-1.5B, Qwen2.5-3B, and DeepSeek-Math-7B. In general, ReSum lifts the average score substantially beyond both GRPO and DGPO baseline. For the smallest Qwen2.5-Math-1.5B, DGPO contributes only 1.32\% over GRPO, while ReSum outperforms DGPO by 2.36\%. This actually implies that the self-summarization incentive alone accounts for an extra 2.36\% on top of difficulty-aware tuning. For Qwen2.5-3B, the average score increases from 25.47\% of GRPO to 28.81\% of ReSum. It can be concluded that the relative improvements are particularly meaningful in smaller models, where limited capacity makes them more vulnerable to forgetting and circular reasoning. ReSum effectively addresses this weakness by teaching the model to compress its own long-range context without external modules. The consistent behavior across backbones suggests that the tree-structured rollout and group-relative advantage design of ReSum capture a wide compatible and model-agnostic benefit.

\begin{table*}[t]
\caption{Ablation results of Artifact  Points (APs), Natural Points (NPs) and our SGPO. Notably w/o SGPO means to disable the dual-group mechanism and use GRPO. The best results are in \textbf{bold}. }
\vspace{-0.1cm}
\centering
\label{tab:my-table:comp} 
\renewcommand{\arraystretch}{1.1}   
\setlength\tabcolsep{6pt} 
\resizebox{1\columnwidth}{!}{
\begin{tabular}{lccccccc}
\toprule[1pt]
Models & AIME24 & AIME25 & AMC23 & MATH500 & Minerva & Olympiad & Avg./$\Delta_{\rm DGPO}$ \\ \hline
Qwen2.5-Math-1.5B & 6.87 & 3.65 & 30.94 & 34.95 & 8.55 & 21.93 & 17.82 \\
GRPO & 11.35 & 3.96 & 46.48 & 64.85 & 20.13 & 29.59 & 29.39 \\
\rowcolor{gray!20}
ReSum & \textbf{15.96} & \textbf{6.83 }& \textbf{51.33} & \textbf{67.12} & \textbf{24.58} & \textbf{32.59} & \textbf{33.07} \\
\,\,w/o APs & 12.50 & 5.10 & 50.41 & 65.40 & 22.43 & 31.41 & 31.21 \\
\,\,w/o NPs & 11.67 & 4.27 & 47.26 & 65.55 & 21.97 & 29.93 & 30.11 \\
\,\,w/o SGPO & 14.64 & 6.52 & 50.59 & 66.13 & 24.30 & 32.15 & 32.39 \\ \hline
Qwen2.5-Math-7B & 12.19 & 4.79 & 35.23 & 48.60 & 15.07 & 16.33 & 22.04 \\
GRPO & 20.94 & 8.44 & 58.98 & 72.20 & 27.76 & 37.33 & 37.61 \\
\rowcolor{gray!20}
ReSum & \textbf{25.42} & 13.33 & \textbf{62.50} & \textbf{76.45} & \textbf{32.44} & \textbf{39.67} & \textbf{41.64} \\
\,\,w/o APs & 24.46 & \textbf{13.48} & 60.39 & 76.00 & 31.03 & 39.01 & 40.73 \\
\,\,w/o NPs & 20.32 & 11.97 & 58.73 & 73.02 & 31.20 & 38.39 & 38.94 \\
\,\,w/o SGPO & 25.11 & 10.56 & 60.91 & 75.59 & 32.02 & 38.66 & 40.48 \\ \bottomrule[1pt]
\end{tabular}}\vspace{-0.4cm}
\end{table*}

\vspace{-0.2cm}
\paragraph{Compatibility with Existing Methods.} Table~\ref{tab:my-table:3} presents the compatibility results of ReSum with orthogonal approaches, where ReSum is applied on top of GPG, DAPO, and GSPO. In every combination, ReSum yields a larger average improvement than augmenting the same base method with DGPO. Starting from 37.93\% for GPG, adding DGPO lifts the score to 38.92\%, whereas adding ReSum achieves 40.76\% (+2.83\%). The pattern holds for DAPO with a +2.94\% gain and for GSPO with a +2.46\% gain. These results highlight that the ability to periodically self-summarize is largely orthogonal to advances in advantage estimation or loss design. While GPG, DAPO, and GSPO each refine how credit is assigned across tokens or groups, they do not actively encourage the model to reset its working context when reasoning becomes redundant. ReSum fills this gap, acting as a general-purpose regularizer that improves reasoning fidelity without interfering with the underlying policy gradient mechanism. This cross-method synergy indicates that summarization behavior can be layered onto future RLVR innovations for cumulative gains.

\vspace{-0.2cm}
\paragraph{Application to Multi-modal Domain.} To study whether ReSum's benefits extend beyond text-only mathematics, we evaluate on the visual question-answering dataset GEOQA-8K with Qwen2.5-3B-VL-Instruct. As shown in Table~\ref{tab:my-table:4}, ReSum reaches an average score of 62.04\%, outperforming GRPO by 5.61\%, while other methods such as DGPO and GPG deliver only modest improvements. Although the multimodal setting introduces additional challenges from visual encoding, the model still benefits from internal summarization of its textual reasoning chain. This suggests that the tendency to produce redundant or self-contradictory statements is not limited to pure mathematical problems but is an inherent property of long-form autoregressive generation across modalities. Crucially, ReSum requires no external compression modules or hand-crafted summary prompts; the model learns to spontaneously determine when to summarize.

\vspace{-0.2cm}
\subsection{Further Analysis} 

\begin{wrapfigure}{r}{0.48\textwidth}
    \centering
     \vspace{-0.3cm}
    \includegraphics[width=1\linewidth]{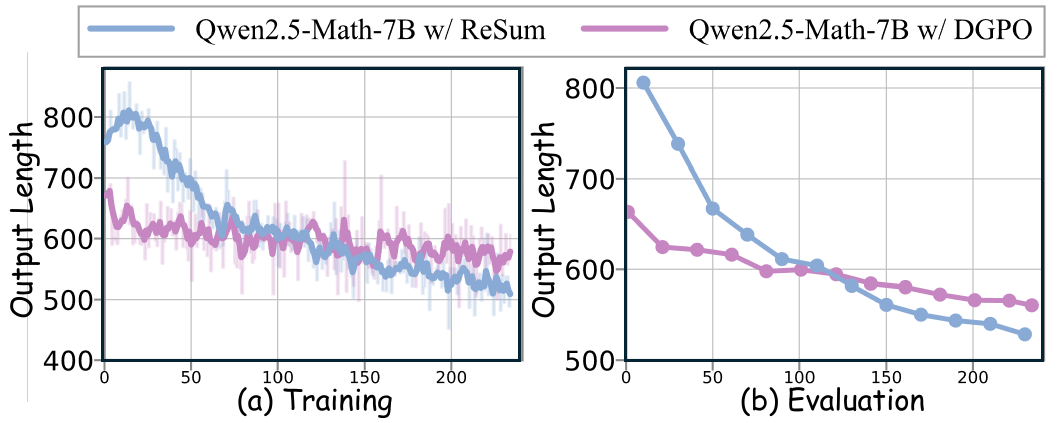} 
    \vspace{-0.5cm}
    \caption{Training and evaluation dynamics of output length with Qwen2.5-Math-7B.} 
    \vspace{-0.6cm}
    \label{fig:len}
\end{wrapfigure}

\vspace{-0.2cm}
\paragraph{Branching Configuration Analysis.} 
We examine how the rollout tree structure affects ReSum under fixed rollout budgets in Table~\ref{tab:my-table:budget}, where the rollout budget is defined as $B=TJ$. 
With a rollout budget of $B=4$, ReSum with $T=4, J=1$ reaches 27.46\%, outperforming the configuration with $T=2, J=2$ by 0.71\%. 
This result suggests that diverse initial paths provide stronger supervision than deeper branching within fewer trees. 
At a rollout budget of $B=16$, ReSum with $T=8, J=2$ achieves 28.55\%, surpassing the 27.98\% obtained by $T=2, J=8$. 
Meanwhile, we find the balanced $T=4, J=4$ configuration yields the highest score among all settings. 
This indicates that combining trajectory diversity with a moderate amount of summarization branching yields the best learning signal. 
At the smallest rollout budget of $B=2$, the minimal tree structure does not yet provide gains, likely because the branching overhead outweighs the benefit when samples are scarce. 

\paragraph{Learning Dynamics.} Figure~\ref{fig:dyna} shows the training and evaluation dynamics of the task reward for ReSum and DGPO, tested on Qwen2.5-Math-1.5B and 7B backbones. For both model scales, ReSum consistently outperforms DGPO across training and inference. While the training-stage improvement may partly reflect latent overfitting effects, the consistently higher upper bound (+0.08 approximately) and faster convergence during inference confirm the effectiveness of ReSum. Its summarization-driven design provides robust and generalizable reasoning enhancements.

We also show the average rollout length during training and evaluation with Qwen2.5-Math-7B in Figure~\ref{fig:len}. Although ReSum initially exhibits an increased rollout length due to the model's unfamiliarity with summarization, it maintains a length reduction throughout the remainder of training in both phases, eventually reducing the average rollout length by approximately 18.6\% relative to DGPO. Notably, this length reduction is accompanied by consistent reward gains, confirming that ReSum's self-summarization mechanism eliminates the typical endless retries in long reasoning chains, rather than simply truncating critical reasoning content to shorten outputs.

\vspace{-0.3cm}
\paragraph{Ablation of Components.} Table~\ref{tab:my-table:comp} reports the ablation results. Removing NPs causes the most severe degradation across both model scales: the average drops sharply from 33.07\% to 30.11\% on the 1.5B model and from 41.64\% to 38.94\% on the 7B model, confirming that natural summarization signals are the primary driver of ReSum's effectiveness by grounding the behavior in authentic reasoning patterns. Excluding APs also yields a notable decline to 31.21\% and 40.73\%, suggesting that artificially injected summarization cues provide complementary supervision. When SGPO is replaced with vanilla GRPO, performance falls to 32.39\% and 40.48\%, respectively. This gap reveals that the tree-based branching alone already supplies useful process-level signals, but the dual-group advantage further refines the learning by rewarding well-placed summaries over poorly positioned ones. Together, APs introduce the summarization signal, NPs expand the summarization  to more fine-grained contexts, and SGPO provides fine-grained credit assignment that distinguishes effective summarization from superficial mimicry. The three components operate synergistically.

\begin{wrapfigure}{r}{0.4\textwidth}
    \centering
     \vspace{-0.3cm}
    \includegraphics[width=1\linewidth]{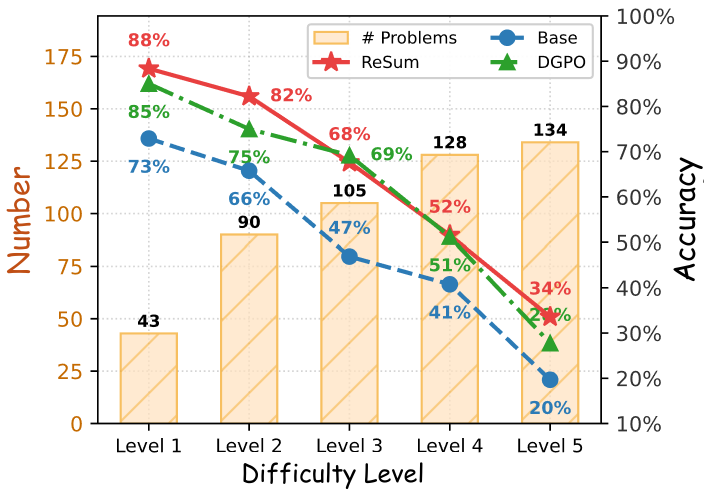} 
    \vspace{-0.5cm}
    \caption{Analysis of accuracy across different difficulty levels of MATH500.} 
    \vspace{-0.6cm}
    \label{fig:ana}
\end{wrapfigure}

\noindent\textbf{Analysis of Accuracy across Problem Difficulty Levels.} Figure~\ref{fig:ana} presents the accuracy  between ReSum and the DGPO baseline across five difficulty levels of MATH500. Higher levels correspond to problems with more intricate logical dependencies and a higher risk of cumulative errors. ReSum delivers consistent accuracy improvements over the baseline at every difficulty level, with the magnitude of the performance gain scaling monotonically with problem complexity. Specifically, ReSum yields a 3\% accuracy lift on the simplest Level 1 problems, where the trend slightly reduces for Levels 2-4. However, ReSum achieves  a 6\% gain on the most challenging Level 5 problems. The reason for this phenomenon is that  although DGPO incorporates the difficulty scaling to foster learning of hard tasks, the model may still fail to effectively learn hard samples due to insufficient capacity. Comparably,  ReSum incentivizes the appropriate self-summarization of LLMs internally, which compresses redundant historical context, preserves core task-relevant information and suppresses error propagation across extended reasoning rollouts.

\vspace{-0.2cm}
\section{Conclusion}
We introduced ReSum, a reinforcement learning framework that incentivizes periodic self-summarization in LLMs, thereby reducing overthinking and context decay without external modules. By expanding rollouts into a tree structure with natural and artificial summarization points, and by applying summarization-aware group-relative advantages, ReSum converts trajectory-level rewards into fine-grained process supervision. Experiments across multiple benchmarks and backbone sizes show consistent accuracy gains of 4\% and an 18.6\% reduction in rollout length, validating that internalized self-summarization is a broadly effective strategy for robust long-chain reasoning.
 
{
    \small
    \bibliographystyle{abbrv}
    \bibliography{main}

@article{dong2025agentic,
  title={Agentic reinforced policy optimization},
  author={Dong, Guanting and Mao, Hangyu and Ma, Kai and Bao, Licheng and Chen, Yifei and Wang, Zhongyuan and Chen, Zhongxia and Du, Jiazhen and Wang, Huiyang and Zhang, Fuzheng and others},
  journal={arXiv preprint arXiv:2507.19849},
  year={2025}
}

@inproceedings{chen2021geoqa,
  title={Geoqa: A geometric question answering benchmark towards multimodal numerical reasoning},
  author={Chen, Jiaqi and Tang, Jianheng and Qin, Jinghui and Liang, Xiaodan and Liu, Lingbo and Xing, Eric and Lin, Liang},
  booktitle={Findings of the Association for Computational Linguistics: ACL-IJCNLP 2021},
  pages={513--523},
  year={2021}
}

@article{wu2025resum,
  title={Resum: Unlocking long-horizon search intelligence via context summarization},
  author={Wu, Xixi and Li, Kuan and Zhao, Yida and Zhang, Liwen and Ou, Litu and Yin, Huifeng and Zhang, Zhongwang and Yu, Xinmiao and Zhang, Dingchu and Jiang, Yong and others},
  journal={arXiv preprint arXiv:2509.13313},
  year={2025}
}

@article{zhao2026training,
  title={Training Multi-Turn Search Agent via Contrastive Dynamic Branch Sampling},
  author={Zhao, Yubao and Huang, Weiquan and Wang, Sudong and Zhao, Ruochen and Chen, Chen and Shu, Yao and Qin, Chengwei},
  journal={arXiv preprint arXiv:2602.03719},
  year={2026}
}

@article{lewkowycz2022solving,
  title={Solving quantitative reasoning problems with language models},
  author={Lewkowycz, Aitor and Andreassen, Anders and Dohan, David and Dyer, Ethan and Michalewski, Henryk and Ramasesh, Vinay and Slone, Ambrose and Anil, Cem and Schlag, Imanol and Gutman-Solo, Theo and others},
  journal={Advances in neural information processing systems},
  volume={35},
  pages={3843--3857},
  year={2022}
}

@article{gao2025beyond,
  title={Beyond ten turns: Unlocking long-horizon agentic search with large-scale asynchronous rl},
  author={Gao, Jiaxuan and Fu, Wei and Xie, Minyang and Xu, Shusheng and He, Chuyi and Mei, Zhiyu and Zhu, Banghua and Wu, Yi},
  journal={arXiv preprint arXiv:2508.07976},
  year={2025}
}

@article{mai2025agent,
  title={Agent rl scaling law: Agent rl with spontaneous code execution for mathematical problem solving},
  author={Mai, Xinji and Xu, Haotian and Li, Zhong-Zhi and Wang, Weinong and Hu, Jian and Zhang, Yingying and Zhang, Wenqiang and others},
  journal={arXiv preprint arXiv:2505.07773},
  year={2025}
}

@article{luo2025gui,
  title={Gui-r1: A generalist r1-style vision-language action model for gui agents},
  author={Luo, Run and Wang, Lu and He, Wanwei and Chen, Longze and Li, Jiaming and Xia, Xiaobo},
  journal={arXiv preprint arXiv:2504.10458},
  year={2025}
}

@article{yuan2025enhancing,
  title={Enhancing visual grounding for gui agents via self-evolutionary reinforcement learning},
  author={Yuan, Xinbin and Zhang, Jian and Li, Kaixin and Cai, Zhuoxuan and Yao, Lujian and Chen, Jie and Wang, Enguang and Hou, Qibin and Chen, Jinwei and Jiang, Peng-Tao and others},
  journal={arXiv preprint arXiv:2505.12370},
  year={2025}
}

@inproceedings{lu2026ui,
  title={Ui-r1: Enhancing efficient action prediction of gui agents by reinforcement learning},
  author={Lu, Zhengxi and Chai, Yuxiang and Guo, Yaxuan and Yin, Xi and Liu, Liang and Wang, Hao and Xiao, Han and Ren, Shuai and Zhao, Pengxiang and Liu, Guangyi and others},
  booktitle={Proceedings of the AAAI Conference on Artificial Intelligence},
  volume={40},
  number={21},
  pages={17608--17616},
  year={2026}
}

@article{shi2025mobilegui,
  title={Mobilegui-rl: Advancing mobile gui agent through reinforcement learning in online environment},
  author={Shi, Yucheng and Yu, Wenhao and Li, Zaitang and Wang, Yonglin and Zhang, Hongming and Liu, Ninghao and Mi, Haitao and Yu, Dong},
  journal={arXiv preprint arXiv:2507.05720},
  year={2025}
}

@article{liu2024rl,
  title={Rl-gpt: Integrating reinforcement learning and code-as-policy},
  author={Liu, Shaoteng and Yuan, Haoqi and Hu, Minda and Li, Yanwei and Chen, Yukang and Liu, Shu and Lu, Zongqing and Jia, Jiaya},
  journal={Advances in Neural Information Processing Systems},
  volume={37},
  pages={28430--28459},
  year={2024}
}

@article{wang2025stepsearch,
  title={Stepsearch: Igniting llms search ability via step-wise proximal policy optimization},
  author={Wang, Ziliang and Zheng, Xuhui and An, Kang and Ouyang, Cijun and Cai, Jialu and Wang, Yuhang and Wu, Yichao},
  journal={arXiv preprint arXiv:2505.15107},
  year={2025}
}

@article{zhang2025agentic,
  title={Agentic context engineering: Evolving contexts for self-improving language models},
  author={Zhang, Qizheng and Hu, Changran and Upasani, Shubhangi and Ma, Boyuan and Hong, Fenglu and Kamanuru, Vamsidhar and Rainton, Jay and Wu, Chen and Ji, Mengmeng and Li, Hanchen and others},
  journal={arXiv preprint arXiv:2510.04618},
  year={2025}
}

@inproceedings{tian2026keepkv,
  title={KeepKV: Achieving Periodic Lossless KV Cache Compression for Efficient LLM Inference},
  author={Tian, Yuxuan and Wang, Zihan and Peng, Yebo and Yuan, Aomufei and Wang, Zhiming and Yi, Bairen and Liu, Xin and Cui, Yong and Yang, Tong},
  booktitle={Proceedings of the AAAI Conference on Artificial Intelligence},
  volume={40},
  number={39},
  pages={33259--33267},
  year={2026}
}

@inproceedings{ma2026compressing,
  title={Compressing kv cache for long-context llm inference with inter-layer attention similarity},
  author={Ma, Da and Chen, Lu and Zhang, Situo and Miao, Yuxun and Zhu, Su and Chen, Zhi and Xu, Hongshen and Li, Hanqi and Fan, Shuai and Pan, Lei and others},
  booktitle={ICASSP 2026-2026 IEEE International Conference on Acoustics, Speech and Signal Processing (ICASSP)},
  pages={16407--16411},
  year={2026},
  organization={IEEE}
}

@article{guo2026less,
  title={When Less is More: The LLM Scaling Paradox in Context Compression},
  author={Guo, Ruishan and Liu, Yibing and Ma, Guoxin and Wang, Yan and Zhang, Yueyang and Xia, Long and Chen, Kecheng and Sun, Zhiyuan and Shi, Daiting},
  journal={arXiv preprint arXiv:2602.09789},
  year={2026}
}

@inproceedings{hou2025treerl,
  title={Treerl: Llm reinforcement learning with on-policy tree search},
  author={Hou, Zhenyu and Hu, Ziniu and Li, Yujiang and Lu, Rui and Tang, Jie and Dong, Yuxiao},
  booktitle={Proceedings of the 63rd Annual Meeting of the Association for Computational Linguistics (Volume 1: Long Papers)},
  pages={12355--12369},
  year={2025}
}

@article{ji2025tree,
  title={Tree search for llm agent reinforcement learning},
  author={Ji, Yuxiang and Ma, Ziyu and Wang, Yong and Chen, Guanhua and Chu, Xiangxiang and Wu, Liaoni},
  journal={arXiv preprint arXiv:2509.21240},
  year={2025}
}

@article{lai2024step,
  title={Step-dpo: Step-wise preference optimization for long-chain reasoning of llms},
  author={Lai, Xin and Tian, Zhuotao and Chen, Yukang and Yang, Senqiao and Peng, Xiangru and Jia, Jiaya},
  journal={arXiv preprint arXiv:2406.18629},
  year={2024}
}

@article{dong2025agentic1,
  title={Agentic entropy-balanced policy optimization},
  author={Dong, Guanting and Bao, Licheng and Wang, Zhongyuan and Zhao, Kangzhi and Li, Xiaoxi and Jin, Jiajie and Yang, Jinghan and Mao, Hangyu and Zhang, Fuzheng and Gai, Kun and others},
  journal={arXiv preprint arXiv:2510.14545},
  year={2025}
}

@article{zhang2025process,
  title={Process vs. outcome reward: Which is better for agentic RAG reinforcement learning},
  author={Zhang, Wenlin and Li, Xiangyang and Dong, Kuicai and Wang, Yichao and Jia, Pengyue and Li, Xiaopeng and Zhang, Yingyi and Xu, Derong and Du, Zhaocheng and Guo, Huifeng and others},
  journal={arXiv preprint arXiv:2505.14069},
  year={2025}
}

@inproceedings{li2025search,
  title={Search-o1: Agentic search-enhanced large reasoning models},
  author={Li, Xiaoxi and Dong, Guanting and Jin, Jiajie and Zhang, Yuyao and Zhou, Yujia and Zhu, Yutao and Zhang, Peitian and Dou, Zhicheng},
  booktitle={Proceedings of the 2025 Conference on Empirical Methods in Natural Language Processing},
  pages={5420--5438},
  year={2025}
}

@article{guo2025deepseek,
  title={Deepseek-r1: Incentivizing reasoning capability in llms via reinforcement learning},
  author={Guo, Daya and Yang, Dejian and Zhang, Haowei and Song, Junxiao and Wang, Peiyi and Zhu, Qihao and Xu, Runxin and Zhang, Ruoyu and Ma, Shirong and Bi, Xiao and others},
  journal={arXiv preprint arXiv:2501.12948},
  year={2025}
}

@article{schulman2017proximal,
  title={Proximal policy optimization algorithms},
  author={Schulman, John and Wolski, Filip and Dhariwal, Prafulla and Radford, Alec and Klimov, Oleg},
  journal={arXiv preprint arXiv:1707.06347},
  year={2017}
}

@article{shao2024deepseekmath,
  title={Deepseekmath: Pushing the limits of mathematical reasoning in open language models},
  author={Shao, Zhihong and Wang, Peiyi and Zhu, Qihao and Xu, Runxin and Song, Junxiao and Bi, Xiao and Zhang, Haowei and Zhang, Mingchuan and Li, YK and Wu, Yang and others},
  journal={arXiv preprint arXiv:2402.03300},
  year={2024}
}

@article{bai1others,
  title={Qwen2. 5-vl technical report},
  author={Bai, Shuai and Chen, Keqin and Liu, Xuejing and Wang, Jialin and Ge, Wenbin and Song, Sibo and Dang, Kai and Wang, Peng and Wang, Shijie and Tang, Jun and others},
  journal={arXiv preprint arXiv:2502.13923},
  volume={4},
  number={5},
  year={1}
}

@inproceedings{liu2026badthink,
  title={Badthink: Triggered overthinking attacks on chain-of-thought reasoning in large language models},
  author={Liu, Shuaitong and Li, Renjue and Yu, Lijia and Zhang, Lijun and Liu, Zhiming and Jin, Gaojie},
  booktitle={Proceedings of the AAAI Conference on Artificial Intelligence},
  volume={40},
  number={38},
  pages={32141--32149},
  year={2026}
}

@article{koh2022empirical,
  title={An empirical survey on long document summarization: Datasets, models, and metrics},
  author={Koh, Huan Yee and Ju, Jiaxin and Liu, Ming and Pan, Shirui},
  journal={ACM computing surveys},
  volume={55},
  number={8},
  pages={1--35},
  year={2022},
  publisher={ACM New York, NY}
}

@article{long2025copy,
  title={Copy-Paste to Mitigate Large Language Model Hallucinations},
  author={Long, Yongchao and Wu, Xian and Zhang, Yingying and Wen, Xianbin and Zhou, Yuxi and Hong, Shenda},
  journal={arXiv preprint arXiv:2510.00508},
  year={2025}
}

@article{ming2024faitheval,
  title={FaithEval: Can Your Language Model Stay Faithful to Context, Even If" The Moon is Made of Marshmallows"},
  author={Ming, Yifei and Purushwalkam, Senthil and Pandit, Shrey and Ke, Zixuan and Nguyen, Xuan-Phi and Xiong, Caiming and Joty, Shafiq},
  journal={arXiv preprint arXiv:2410.03727},
  year={2024}
}

@inproceedings{longpre2021entity,
  title={Entity-based knowledge conflicts in question answering},
  author={Longpre, Shayne and Perisetla, Kartik and Chen, Anthony and Ramesh, Nikhil and DuBois, Chris and Singh, Sameer},
  booktitle={Proceedings of the 2021 conference on empirical methods in natural language processing},
  pages={7052--7063},
  year={2021}
}

@article{srivastava2025llms,
  title={Do LLMs Overthink Basic Math Reasoning? Benchmarking the Accuracy-Efficiency Tradeoff in Language Models},
  author={Srivastava, Gaurav and Hussain, Aafiya and Srinivasan, Sriram and Wang, Xuan},
  journal={arXiv preprint arXiv:2507.04023},
  year={2025}
}

@article{wu2025more,
  title={When more is less: Understanding chain-of-thought length in llms},
  author={Wu, Yuyang and Wang, Yifei and Ye, Ziyu and Du, Tianqi and Jegelka, Stefanie and Wang, Yisen},
  journal={arXiv preprint arXiv:2502.07266},
  year={2025}
}

@article{chen2024not,
  title={Do not think that much for 2+ 3=? on the overthinking of o1-like llms},
  author={Chen, Xingyu and Xu, Jiahao and Liang, Tian and He, Zhiwei and Pang, Jianhui and Yu, Dian and Song, Linfeng and Liu, Qiuzhi and Zhou, Mengfei and Zhang, Zhuosheng and others},
  journal={arXiv preprint arXiv:2412.21187},
  year={2024}
}

@article{zhang2025llms,
  title={Do LLMs Really Need 10+ Thoughts for" Find the Time 1000 Days Later"? Towards Structural Understanding of LLM Overthinking},
  author={Zhang, Xinliang Frederick and Mohananey, Anhad and Chronopoulou, Alexandra and Papalampidi, Pinelopi and Gupta, Somit and Munkhdalai, Tsendsuren and Wang, Lu and Upadhyay, Shyam},
  journal={arXiv preprint arXiv:2510.07880},
  year={2025}
}

@inproceedings{zhang2025grpo,
  title={Grpo-lead: A difficulty-aware reinforcement learning approach for concise mathematical reasoning in language models},
  author={Zhang, Jixiao and Zuo, Chunsheng},
  booktitle={Proceedings of the 2025 Conference on Empirical Methods in Natural Language Processing},
  pages={5642--5665},
  year={2025}
}

@misc{chen2025vinci,
  title={Vinci. R1-v: Reinforcing super generalization ability in vision-language models with less than \$3},
  author={Chen, Liang and Li, Lei and Zhao, Haozhe and Song, Yifan},
  year={2025}
}

@techreport{yang2412qiu,
  title={Qwen2. 5 technical report},
  author={Yang, A and Yang, B and Zhang, B and Hui, B and Zheng, B and Yu, B and Li, C and Liu, D and Huang, F and Wei, H and others},
  year={2024}
}

@article{feng2025group,
  title={Group-in-group policy optimization for llm agent training},
  author={Feng, Lang and Xue, Zhenghai and Liu, Tingcong and An, Bo},
  journal={arXiv preprint arXiv:2505.10978},
  year={2025}
}

@article{ma2026skillclaw,
  title={SkillClaw: Let Skills Evolve Collectively with Agentic Evolver},
  author={Ma, Ziyu and Yang, Shidong and Ji, Yuxiang and Wang, Xucong and Wang, Yong and Hu, Yiming and Huang, Tongwen and Chu, Xiangxiang},
  journal={arXiv preprint arXiv:2604.08377},
  year={2026}
}

@inproceedings{chu2025gpg,
title={{GPG}: A Simple and Strong Reinforcement Learning Baseline for Model Reasoning},
author={Xiangxiang Chu and Hailang Huang and Xiao Zhang and Fei Wei and Yong Wang},
booktitle={The Fourteenth International Conference on Learning Representations},
year={2026},
url={https://openreview.net/forum?id=inccdtfx8x}
}

@article{zhao2025geometric,
  title={Geometric-mean policy optimization},
  author={Zhao, Yuzhong and Liu, Yue and Liu, Junpeng and Chen, Jingye and Wu, Xun and Hao, Yaru and Lv, Tengchao and Huang, Shaohan and Cui, Lei and Ye, Qixiang and others},
  journal={arXiv preprint arXiv:2507.20673},
  year={2025}
}

@inproceedings{li2026adacurl,
  title={Adacurl: Adaptive curriculum reinforcement learning with invalid sample mitigation and historical revisiting},
  author={Li, Renda and Huang, Hailang and Wei, Fei and Xiong, Feng and Wang, Yong and Chu, Xiangxiang},
  booktitle={Proceedings of the AAAI Conference on Artificial Intelligence},
  volume={40},
  number={27},
  pages={23123--23131},
  year={2026}
}

@inproceedings{
dai2026harder,
title={Harder Is Better: Boosting Mathematical Reasoning via Difficulty-Aware {GRPO} and Multi-Aspect Question Reformulation},
author={Yanqi Dai and Yuxiang Ji and Xiao Zhang and Yong Wang and Xiangxiang Chu and Zhiwu Lu},
booktitle={The Fourteenth International Conference on Learning Representations},
year={2026},
url={https://openreview.net/forum?id=nfURupkdRJ}
}

@inproceedings{dai2025pretraining,
  title={Pretraining context compressor for large language models with embedding-based memory},
  author={Dai, Yuhong and Lian, Jianxun and Huang, Yitian and Zhang, Wei and Zhou, Mingyang and Wu, Mingqi and Xie, Xing and Liao, Hao},
  booktitle={Proceedings of the 63rd Annual Meeting of the Association for Computational Linguistics (Volume 1: Long Papers)},
  pages={28715--28732},
  year={2025}
}

@article{ge2023context,
  title={In-context autoencoder for context compression in a large language model},
  author={Ge, Tao and Hu, Jing and Wang, Lei and Wang, Xun and Chen, Si-Qing and Wei, Furu},
  journal={arXiv preprint arXiv:2307.06945},
  year={2023}
}

@article{zheng2025group,
  title={Group sequence policy optimization},
  author={Zheng, Chujie and Liu, Shixuan and Li, Mingze and Chen, Xiong-Hui and Yu, Bowen and Gao, Chang and Dang, Kai and Liu, Yuqiong and Men, Rui and Yang, An and others},
  journal={arXiv preprint arXiv:2507.18071},
  year={2025}
}

@inproceedings{chen2024long,
  title={Long context is not long at all: A prospector of long-dependency data for large language models},
  author={Chen, Longze and Liu, Ziqiang and He, Wanwei and Zheng, Yinhe and Sun, Hao and Li, Yunshui and Luo, Run and Yang, Min},
  booktitle={Proceedings of the 62nd Annual Meeting of the Association for Computational Linguistics (Volume 1: Long Papers)},
  pages={8222--8234},
  year={2024}
}

@article{face2025open,
  title={Open r1: A fully open reproduction of deepseek-r1, January 2025},
  author={Face, Hugging},
  journal={URL https://github. com/huggingface/open-r1},
  volume={7},
  year={2025}
}

@inproceedings{yang2024pyramidinfer,
  title={Pyramidinfer: Pyramid kv cache compression for high-throughput llm inference},
  author={Yang, Dongjie and Han, XiaoDong and Gao, Yan and Hu, Yao and Zhang, Shilin and Zhao, Hai},
  booktitle={Findings of the Association for Computational Linguistics: ACL 2024},
  pages={3258--3270},
  year={2024}
}

@article{ling2025longreason,
  title={Longreason: A synthetic long-context reasoning benchmark via context expansion},
  author={Ling, Zhan and Liu, Kang and Yan, Kai and Yang, Yifan and Lin, Weijian and Fan, Ting-Han and Shen, Lingfeng and Du, Zhengyin and Chen, Jiecao},
  journal={arXiv preprint arXiv:2501.15089},
  year={2025}
}

@inproceedings{dong2025longred,
  title={Longred: Mitigating short-text degradation of long-context large language models via restoration distillation},
  author={Dong, Zican and Li, Junyi and Jiang, Jinhao and Xu, Mingyu and Zhao, Wayne Xin and Wang, Bingning and Chen, Weipeng},
  booktitle={Proceedings of the 63rd Annual Meeting of the Association for Computational Linguistics (Volume 1: Long Papers)},
  pages={10687--10707},
  year={2025}
}

@inproceedings{liu2024forgetting,
  title={Forgetting curve: A reliable method for evaluating memorization capability for long-context models},
  author={Liu, Xinyu and Zhao, Runsong and Huang, Pengcheng and Xiao, Chunyang and Li, Bei and Wang, Jingang and Xiao, Tong and Zhu, Jingbo},
  booktitle={Proceedings of the 2024 Conference on Empirical Methods in Natural Language Processing},
  pages={4667--4682},
  year={2024}
}

@article{yu2025dapo,
  title={Dapo: An open-source llm reinforcement learning system at scale},
  author={Yu, Qiying and Zhang, Zheng and Zhu, Ruofei and Yuan, Yufeng and Zuo, Xiaochen and Yue, Yu and Dai, Weinan and Fan, Tiantian and Liu, Gaohong and Liu, Lingjun and others},
  journal={arXiv preprint arXiv:2503.14476},
  year={2025}
}

@article{liu2025understanding,
  title={Understanding r1-zero-like training: A critical perspective},
  author={Liu, Zichen and Chen, Changyu and Li, Wenjun and Qi, Penghui and Pang, Tianyu and Du, Chao and Lee, Wee Sun and Lin, Min},
  journal={arXiv preprint arXiv:2503.20783},
  year={2025}
}

@article{xie2024monte,
  title={Monte carlo tree search boosts reasoning via iterative preference learning},
  author={Xie, Yuxi and Goyal, Anirudh and Zheng, Wenyue and Kan, Min-Yen and Lillicrap, Timothy P and Kawaguchi, Kenji and Shieh, Michael},
  journal={arXiv preprint arXiv:2405.00451},
  year={2024}
}

@article{ma2026fipo,
  title={FIPO: Eliciting Deep Reasoning with Future-KL Influenced Policy Optimization},
  author={Ma, Chiyu and Yang, Shuo and Huang, Kexin and Lu, Jinda and Meng, Haoming and Wang, Shangshang and Ding, Bolin and Vosoughi, Soroush and Wang, Guoyin and Zhou, Jingren},
  journal={arXiv preprint arXiv:2603.19835},
  year={2026}
}

@inproceedings{paul2024refiner,
  title={Refiner: Reasoning feedback on intermediate representations},
  author={Paul, Debjit and Ismayilzada, Mete and Peyrard, Maxime and Borges, Beatriz and Bosselut, Antoine and West, Robert and Faltings, Boi},
  booktitle={Proceedings of the 18th Conference of the European Chapter of the Association for Computational Linguistics (Volume 1: Long Papers)},
  pages={1100--1126},
  year={2024}
}

@inproceedings{hooper2025squeezed,
  title={Squeezed attention: Accelerating long context length llm inference},
  author={Hooper, Coleman Richard Charles and Kim, Sehoon and Mohammadzadeh, Hiva and Maheswaran, Monishwaran and Zhao, Sebastian and Paik, June and Mahoney, Michael W and Keutzer, Kurt and Gholami, Amir},
  booktitle={Proceedings of the 63rd Annual Meeting of the Association for Computational Linguistics (Volume 1: Long Papers)},
  pages={32631--32652},
  year={2025}
}

@inproceedings{muennighoff2025s1,
  title={s1: Simple test-time scaling},
  author={Muennighoff, Niklas and Yang, Zitong and Shi, Weijia and Li, Xiang Lisa and Fei-Fei, Li and Hajishirzi, Hannaneh and Zettlemoyer, Luke and Liang, Percy and Cand{\`e}s, Emmanuel and Hashimoto, Tatsunori B},
  booktitle={Proceedings of the 2025 Conference on Empirical Methods in Natural Language Processing},
  pages={20286--20332},
  year={2025}
}

@inproceedings{li2025reflectevo,
  title={Reflectevo: Improving meta introspection of small llms by learning self-reflection},
  author={Li, Jiaqi and Dong, Xinyi and Liu, Yang and Yang, Zhizhuo and Wang, Quansen and Wang, Xiaobo and Zhu, Song-Chun and Jia, Zixia and Zheng, Zilong},
  booktitle={Findings of the Association for Computational Linguistics: ACL 2025},
  pages={16948--16966},
  year={2025}
}

@article{bensal2025reflect,
  title={Reflect, retry, reward: Self-improving llms via reinforcement learning},
  author={Bensal, Shelly and Jamil, Umar and Bryant, Christopher and Russak, Melisa and Kamble, Kiran and Mozolevskyi, Dmytro and Ali, Muayad and AlShikh, Waseem},
  journal={arXiv preprint arXiv:2505.24726},
  year={2025}
}

@inproceedings{li2025encouraging,
  title={Encouraging good processes without the need for good answers: Reinforcement learning for llm agent planning},
  author={Li, Zhiwei and Hu, Yong and Wang, Wenqing},
  booktitle={Proceedings of the 2025 Conference on Empirical Methods in Natural Language Processing: Industry Track},
  pages={1654--1666},
  year={2025}
}

@article{berton2025compllm,
  title={CompLLM: Compression for Long Context Q\&A},
  author={Berton, Gabriele and Unnikrishnan, Jayakrishnan and Tran, Son and Shah, Mubarak},
  journal={arXiv preprint arXiv:2509.19228},
  year={2025}
}

@article{zhao2025boosting,
  title={Boosting llm reasoning via spontaneous self-correction},
  author={Zhao, Xutong and Xu, Tengyu and Wang, Xuewei and Chen, Zhengxing and Jin, Di and Tan, Liang and Yu, Zishun and Zhao, Zhuokai and He, Yun and Wang, Sinong and others},
  journal={arXiv preprint arXiv:2506.06923},
  year={2025}
}

@article{xiong2025self,
  title={Self-rewarding correction for mathematical reasoning},
  author={Xiong, Wei and Zhang, Hanning and Ye, Chenlu and Chen, Lichang and Jiang, Nan and Zhang, Tong},
  journal={arXiv preprint arXiv:2502.19613},
  year={2025}
}

@article{petrov2025long,
  title={Long context in-context compression by getting to the gist of gisting},
  author={Petrov, Aleksandar and Sandler, Mark and Zhmoginov, Andrey and Miller, Nolan and Vladymyrov, Max},
  journal={arXiv preprint arXiv:2504.08934},
  year={2025}
}

@article{li2025upfront,
  title={Upfront Chain-of-Thought: A Cooperative Framework for Chain-of-Thought Compression},
  author={Li, Chengzhengxu and Liu, Xiaoming and Zhang, Zhaohan and Zhang, Shaochu and Liu, Shengchao and Ma, Guoxin and Lan, Yu and Shen, Chao},
  journal={arXiv preprint arXiv:2510.08647},
  year={2025}
}

@article{wang2026textbf,
  title={\textbf{Re}$^2$: Unlocking LLM Reasoning via Reinforcement Learning with Re-solving},
  author={Wang, Pinzheng and Xu, Shuli and Li, Juntao and Luo, Yu and Li, Dong and Hao, Jianye and Zhang, Min},
  journal={arXiv preprint arXiv:2603.07197},
  year={2026}
}

@article{rafailov2023direct,
  title={Direct preference optimization: Your language model is secretly a reward model},
  author={Rafailov, Rafael and Sharma, Archit and Mitchell, Eric and Manning, Christopher D and Ermon, Stefano and Finn, Chelsea},
  journal={Advances in neural information processing systems},
  volume={36},
  pages={53728--53741},
  year={2023}
}

@article{hendrycks2021measuring,
  title={Measuring mathematical problem solving with the math dataset},
  author={Hendrycks, Dan and Burns, Collin and Kadavath, Saurav and Arora, Akul and Basart, Steven and Tang, Eric and Song, Dawn and Steinhardt, Jacob},
  journal={arXiv preprint arXiv:2103.03874},
  year={2021}
}

@inproceedings{lightman2023let,
  title={Let's verify step by step},
  author={Lightman, Hunter and Kosaraju, Vineet and Burda, Yuri and Edwards, Harrison and Baker, Bowen and Lee, Teddy and Leike, Jan and Schulman, John and Sutskever, Ilya and Cobbe, Karl},
  booktitle={The twelfth international conference on learning representations},
  year={2023}
}

@article{yang2024qwen2,
  title={Qwen2. 5-math technical report: Toward mathematical expert model via self-improvement},
  author={Yang, An and Zhang, Beichen and Hui, Binyuan and Gao, Bofei and Yu, Bowen and Li, Chengpeng and Liu, Dayiheng and Tu, Jianhong and Zhou, Jingren and Lin, Junyang and others},
  journal={arXiv preprint arXiv:2409.12122},
  year={2024}
}

@inproceedings{he2024olympiadbench,
  title={Olympiadbench: A challenging benchmark for promoting agi with olympiad-level bilingual multimodal scientific problems},
  author={He, Chaoqun and Luo, Renjie and Bai, Yuzhuo and Hu, Shengding and Thai, Zhen and Shen, Junhao and Hu, Jinyi and Han, Xu and Huang, Yujie and Zhang, Yuxiang and others},
  booktitle={Proceedings of the 62nd Annual Meeting of the Association for Computational Linguistics (Volume 1: Long Papers)},
  pages={3828--3850},
  year={2024}
}

@misc{aime25,
      title={American Invitational Mathematics Examination (AIME) 2025}, 
      author={Zhang, Yifan and Math-AI, Team},
      year={2025},
}

@misc{aime24,
      title={American Invitational Mathematics Examination (AIME) 2024}, 
      author={Zhang, Yifan and Math-AI, Team},
      year={2024},
}

@inproceedings{wang2023plan,
  title={Plan-and-solve prompting: Improving zero-shot chain-of-thought reasoning by large language models},
  author={Wang, Lei and Xu, Wanyu and Lan, Yihuai and Hu, Zhiqiang and Lan, Yunshi and Lee, Roy Ka-Wei and Lim, Ee-Peng},
  booktitle={Proceedings of the 61st annual meeting of the association for computational linguistics (volume 1: long papers)},
  pages={2609--2634},
  year={2023}
}

@article{jin2025search,
  title={Search-r1: Training llms to reason and leverage search engines with reinforcement learning},
  author={Jin, Bowen and Zeng, Hansi and Yue, Zhenrui and Yoon, Jinsung and Arik, Sercan and Wang, Dong and Zamani, Hamed and Han, Jiawei},
  journal={arXiv preprint arXiv:2503.09516},
  year={2025}
}
}

\clearpage
\newpage
\appendix

 \begin{center}\Large{\textbf{\appendixname{\\\texttt{ReSum}: Synergizing LLM Reasoning and Summarization with Reinforcement Learning}}}\end{center}
The Appendix of the paper is organized as:
\begin{itemize}
    \item Appendix \ref{appx:a}: We give the proof of the proposition.
    \item Appendix \ref{appx:b}: We introduce combinations of ReSum and other methods.
    \item Appendix \ref{appx:c}: We report datasets and baselines.
    \item Appendix \ref{appx:d}: We report implementation details.
    \item Appendix \ref{appx:e}: We report the further analysis.
    \item Appendix \ref{appx:f}: We report the algorithm.
    \item Appendix \ref{appx:g}: We report the impact statement.
    \item Appendix \ref{appx:h}: We report the limitations.
    \item Appendix \ref{appx:i}: We report the use of LLMs. 
\end{itemize}

\section{Proof of Proposition \ref{prop:1}}\label{appx:a}
 
We prove the proposition in two steps: first we show that a $\delta$-faithful summary induces only a small distribution shift; then we use the local stability of the reward to conclude that the expected return of the summarized branch is not worse than that of the non-summarized branch, up to a small error term.

According to our assumptions in the proposition, the summary operator $\mathcal{S}$ is $\delta$-faithful, i.e.,
$\Delta_i=\mathbb{D}_{\mathrm{KL}}\!\left(p^\star(\cdot \mid q, H_{<i}) \,\|\, p^\star(\cdot \mid q, \mathcal{S}(H_{<i}))\right)\le \delta.$
Let $\pi_{\mathrm{full}}(\cdot) = p^\star(\cdot \mid q, H_{<i})$ and $\pi_{\mathrm{sum}}(\cdot) = p^\star(\cdot \mid q, \mathcal{S}(H_{<i})).$
Then by Pinsker's inequality,
$$
D_{\mathrm{TV}}(\pi_{\mathrm{full}}, \pi_{\mathrm{sum}})
\le
\sqrt{\frac{1}{2}\mathbb{D}_{\mathrm{KL}}(\pi_{\mathrm{full}} \| \pi_{\mathrm{sum}})}
\le
\sqrt{\frac{\delta}{2}}.
$$
So the summary branch remains close to the full-history branch in total variation distance. Then, we assume the reward is locally stable with respect to the continuation distribution, meaning that there exists a constant $L>0$ such that for any two continuation policies $\pi_1,\pi_2$,
$$
\left|\mathcal{R}(\pi_1)-\mathcal{R}(\pi_2)\right|
\le
L \cdot D_{\mathrm{TV}}(\pi_1,\pi_2).
$$
Applying this to $\pi_{\mathrm{full}}$ and $\pi_{\mathrm{sum}}$, we obtain
$$
\left|
\mathcal{R}(\pi_{\mathrm{full}})-\mathcal{R}(\pi_{\mathrm{sum}})
\right|
\le
L \cdot D_{\mathrm{TV}}(\pi_{\mathrm{full}}, \pi_{\mathrm{sum}})
\le
L\sqrt{\frac{\delta}{2}},
$$
which implies
$$
\mathcal{R}(\pi_{\mathrm{sum}})
\ge
\mathcal{R}(\pi_{\mathrm{full}}) - L\sqrt{\frac{\delta}{2}}.
$$ 
Now we have $R_{\mathrm{nosum}}(i) = \mathcal{R}(\pi_{\mathrm{full}})$ and
$R_{\mathrm{sum}}(i) = \mathcal{R}(\pi_{\mathrm{sum}})$, which yields
$$
G_i
=
R_{\mathrm{sum}}(i)-R_{\mathrm{nosum}}(i)
\ge
-L\sqrt{\frac{\delta}{2}}.
$$
This proves the first part of the proposition. Also, the bound above shows that faithful summarization cannot hurt too much. To obtain strict improvement, we further observe that the summary operation removes redundant context. Let the full prefix be decomposed into task-relevant information and redundant noise: $H_{<i} = H^{\mathrm{rel}}_{<i} \cup H^{\mathrm{red}}_{<i}.$

A faithful summary approximately preserves $H^{\mathrm{rel}}_{<i}$ while shrinking or discarding $H^{\mathrm{red}}_{<i}$. Since redundant history increases the chance of drifting, repeating, or overthinking, removing it decreases the effective uncertainty in future generation. Therefore, whenever the reward degradation induced by compression is smaller than the gain from reduced noise, we have
$R_{\mathrm{sum}}(i) > R_{\mathrm{nosum}}(i)$, which means $G_i>0$.
 
In summary, combining the above arguments, a $\delta$-faithful summary induces only a small distribution shift, and under local reward stability this yields a non-negative reward gap up to a small error term. When the removed context is mostly redundant, the summary branch can strictly outperform the non-summary branch. This establishes the proposition. 

\section{Details of Combining ReSum and Other Methods}\label{appx:b}

This section introduces how ReSum is integrated with other policy optimization methods. Since ReSum primarily modifies the advantage estimation and introduces a tree-structured rollout, the combination is straightforward: we keep the original loss formulation of each baseline method and replace their advantage terms with our summarization-aware group-relative advantage defined in Eq.~\eqref{eq:sumnosum}. The following subsections provide the exact optimization objectives for each combination.

\subsection{Combination with GPG}

GPG removes the clipping and KL-regularization terms from GRPO and directly optimizes the policy gradient. When combined with ReSum, we retain GPG's unclipped objective while replacing its original group-relative advantage with our summarization-aware dual-level advantage. For a batch of queries $\{q_s\}_{s=1}^{S}$ and their corresponding rollout trees $\{\mathcal{T}_{s,t}\}_{t=1}^{T}$, the optimization objective is:

\begin{equation}
\begin{split}
    \mathcal{J}_{\mathrm{GPG + ReSum}}(\theta) 
    =& \mathbb{E}_{\{q_s\}_{s=1}^S \sim \mathcal{D}, \{\mathcal{T}_{s,t}\}_{t=1}^T \sim \pi_{\theta_{\mathrm{old}}}(\cdot \mid q_s)}
    \Bigg[
    \frac{1}{\sum_{s=1}^{S} \sum_{t=1}^{T} |\mathcal{T}_{s,t}|}  \\
    &\sum_{s=1}^{S} \sum_{t=1}^{T} \sum_{e=1}^{|\mathcal{T}_{s,t}|}
    \frac{1}{|\tau_{s,t,e}|}
    \sum_{i=1}^{|\tau_{s,t,e}|}
    \left[-\log \pi_{\theta}(\tau_{s,t,e,i} \mid q_s, \tau_{s,t,e,<i}) \, \hat{A}_{s,t,e,i}\right]
    \Bigg],
\end{split}
\end{equation}

where $\hat{A}_{s,t,e,i}$ is the summarization-aware group-relative advantage computed by Eq.~\eqref{eq:sumnosum}, combining both $\hat{A}^{\mathrm{sum}}$ and $\hat{A}^{\mathrm{nosum}}$ advantages over the rollout tree.

\subsection{Combination with DAPO}

DAPO employs a PPO-style clipped objective with asymmetric clipping bounds and a composite reward signal that includes both accuracy and length penalty (i.e., $r_{s,t,e} = r_{\mathrm{acc},s,t,e} + r_{\mathrm{length},s,t,e}$). When integrating ReSum with DAPO, we preserve DAPO's clipped objective, token-level loss computation, and overlong reward shaping, while substituting its advantage estimation with our summarization-aware group-relative advantage. Notably, the composite reward is used for advantage calculation in Eq.~\eqref{eq:reward} and Eq.~\eqref{eq:sumnosum}. The optimization objective is:

\begin{equation}
\begin{aligned}
\mathcal{J}_{\mathrm{DAPO + ReSum}}(\theta) 
=& \mathbb{E}_{\{q_s\}_{s=1}^S \sim \mathcal{D}, \{\mathcal{T}_{s,t}\}_{t=1}^T \sim \pi_{\theta_{\mathrm{old}}}(\cdot \mid q_s)}
\Bigg[
\frac{1}{\sum_{s=1}^{S} \sum_{t=1}^{T} |\mathcal{T}_{s,t}|} \\
& \sum_{s=1}^{S} \sum_{t=1}^{T} \sum_{e=1}^{|\mathcal{T}_{s,t}|}
\frac{1}{|\tau_{s,t,e}|}
\sum_{i=1}^{|\tau_{s,t,e}|}
\min \Big(\rho_{s,t,e,i}(\theta) \hat{A}_{s,t,e,i}, \\
& \mathrm{clip}(\rho_{s,t,e,i}(\theta), 1 - \epsilon_{\mathrm{low}}, 1 + \epsilon_{\mathrm{high}}) \hat{A}_{s,t,e,i}\Big)
\Bigg],
\end{aligned}
\end{equation}

where $\rho_{s,t,e,i}(\theta)$ is the token-level importance sampling ratio defined in Eq.~\eqref{eq:obj}, and $\hat{A}_{s,t,e,i}$ is our summarization-aware group-relative advantage. The asymmetric clipping bounds $\epsilon_{\mathrm{low}}$ and $\epsilon_{\mathrm{high}}$ follow the original DAPO design.

\subsection{Combination with GSPO}

GSPO operates at the sequence level, computing a sequence-level importance sampling ratio and applying PPO-style clipping over entire responses. To integrate ReSum with GSPO, we retain GSPO's sequence-level formulation and replace its group-relative advantage with our summarization-aware dual-level advantage. The optimization objective is:

\begin{equation}
\begin{aligned}
\mathcal{J}_{\mathrm{GSPO + ReSum}}(\theta) 
=& \mathbb{E}_{\{q_s\}_{s=1}^S \sim \mathcal{D}, \{\mathcal{T}_{s,t}\}_{t=1}^T \sim \pi_{\theta_{\mathrm{old}}}(\cdot \mid q_s)}
\Bigg[
\frac{1}{\sum_{s=1}^{S} \sum_{t=1}^{T} |\mathcal{T}_{s,t}|}
\sum_{s=1}^{S} \sum_{t=1}^{T} \sum_{e=1}^{|\mathcal{T}_{s,t}|} \\
&  \min \Big(S_{s,t,e}(\theta) \hat{A}_{s,t,e}, \mathrm{clip}(S_{s,t,e}(\theta), 1 - \epsilon, 1 + \epsilon) \hat{A}_{s,t,e}\Big)
\Bigg], \\
&S_{s,t,e}(\theta) = \left(\prod_{i=1}^{|\tau_{s,t,e}|} \rho_{s,t,e,i}(\theta)\right)^{1/|\tau_{s,t,e}|}.
\end{aligned}
\end{equation}

where $S_{s,t,e}(\theta)$ is the sequence-level importance sampling ratio for the $e$-th rollout in the $t$-th tree. Here $\hat{A}_{s,t,e}$ denotes the sequence-level summarization-aware advantage obtained by averaging our token-level dual-group advantage $\hat{A}_{s,t,e,i}$ over the entire rollout. This combination allows GSPO to benefit from ReSum's fine-grained process supervision while maintaining its sequence-level update granularity.

In all the above combinations, the total rollout budget per query remains identical to the original baselines. The tree-structured expansion of ReSum is applied before advantage computation, and the group partitioning into $G_{\mathrm{sum}}$ and $G_{\mathrm{nosum}}$ follows the same procedure described in Section \ref{sec:33} of the main paper.
\section{Datasets and  Baselines}\label{appx:c} 
The datasets are introduced as follows:

\noindent\textbf{MATH.} The MATH dataset~\cite{hendrycks2021measuring} provides 12,500 competition-level problems drawn from high school contests such as AMC-10, AMC-12, and AIME, each labeled with difficulty levels 1-5,  and spanning seven subject areas including algebra, geometry, and number theory. Following DGPO~\cite{dai2026harder}, we use the 7,500-problem training split for RLVR training.

\noindent\textbf{AIME24} AIME24 is collected from the two 2024 American Invitational Mathematics Examination contests. It contains 30 open-ended problems with integer answers between 0 and 999.  AIME24 has become a widely adopted stress test for advanced mathematical reasoning in LLMs, and we report scores averaged over 32 evaluation runs to ensure reliable comparisons.

\noindent\textbf{AIME25.} Similarly, AIME25 is collected from the two 2025 American Invitational Mathematics Examination contests. It features 30 integer-answer problems from AIME-I and AIME-II. Because it is the most recent release, AIME25 serves as a clean evaluation set that guards against potential data contamination in older benchmarks, making it a stringent probe of genuine problem-solving ability. We also average results over 32 runs for this benchmark.

\noindent\textbf{AMC23} AMC23 offers 46 multiple-choice problems covering algebra, geometry, number theory, and combinatorics at the high school level. As the qualifying stage for the AIME, AMC problems are somewhat more accessible than AIME tasks but still require solid mathematical reasoning. We evaluate on AMC23 with 32-run averaging.

\noindent\textbf{MATH500.} MATH500~\cite{lightman2023let} is a 500-problem subset of the MATH test set, uniformly sampled to retain the original distribution across difficulty levels and subjects. Introduced by OpenAI for process-supervised reasoning studies, it has become the standard in-distribution benchmark for models trained on MATH. Evaluation follows the convention of exact match on answers extracted from \verb|\boxed{}|, and we report averages over 4 runs.

\noindent\textbf{Minerva.} Minerva~\cite{lewkowycz2022solving} draws its 272 problems primarily from OpenCourseWare materials, spanning university and graduate-level STEM subjects such as astronomy, solid-state chemistry, differential equations, and special relativity. Rather than competition mathematics, Minerva tests the model's ability to apply quantitative reasoning in scientific contexts, complementing the competition-focused benchmarks with a broader probe of scientific problem solving. Results are averaged over 4 runs.

\noindent\textbf{OlympiadBench.} OlympiadBench~\cite{he2024olympiadbench} assembles 8,476 expert-level problems from international and Chinese Olympiad competitions in mathematics and physics, together with items from the Chinese college entrance exam (Gaokao). It offers bilingual (English/Chinese) and multimodal problem formats with detailed step-by-step annotations, making it one of the most demanding tests of cross-lingual, multi-domain mathematical reasoning. We report 4-run averages on this benchmark.

\noindent\textbf{GEOQA-8K.} GEOQA-8K~\cite{chen2025vinci} is an extended version of the GeoQA geometric question answering corpus~\cite{chen2021geoqa}. It contains 8,000 problems that pair textual descriptions with visual diagrams, requiring joint reasoning over language and geometry. Each problem includes an annotated solving program that makes the reasoning chain explicit. We use GEOQA-8K to test whether ReSum's self-summarization gains transfer to multimodal settings, where the additional visual modality increases context complexity. Results are averaged over 4 evaluation runs.

The baselines are introduced as follows:

GRPO \cite{guo2025deepseek} is a critic-free reinforcement learning algorithm that eliminates the value network by sampling a group of candidate responses per prompt and computing normalized group-relative advantages as training signals. Originally developed for DeepSeek's mathematical reasoning pipeline, GRPO significantly reduces memory and computational costs compared to PPO while maintaining competitive training stability.

Dr. GRPO \cite{liu2025understanding} critically examines the statistical biases embedded in the standard GRPO objective, identifying that the per-sample normalization scheme introduces systematic gradient distortions that can impede learning. It proposes a debiased reformulation of the advantage estimator that corrects these biases, offering a theoretically grounded perspective on R1-Zero-like training and improving optimization reliability without additional computational overhead.

GPG \cite{chu2025gpg} revisits the classical policy gradient mechanism and proposes a minimalist reinforcement learning approach called Group Policy Gradient that directly optimizes the original RL objective without relying on surrogate loss functions or clipping. By stripping away the approximations introduced by PPO-style methods, GPG demonstrates that a clean, unmodified policy gradient formulation can serve as a surprisingly strong and competitive baseline for LLM reasoning tasks.

DAPO \cite{yu2025dapo} is an open-source, industrial-scale RL training system for LLMs that introduces four targeted algorithmic improvements over GRPO: Clip-Higher for asymmetric clipping to encourage output diversity, dynamic sampling to filter uninformative training instances, token-level policy gradient loss for more granular optimization, and overlong reward shaping to penalize excessively verbose responses. Together, these techniques address the entropy collapse and reward hacking issues that commonly arise in large-scale chain-of-thought RL training.

GSPO \cite{zheng2025group} replaces the token-level importance ratio used in PPO and GRPO with a sequence-level importance ratio, arguing that sequence-level clipping provides a more principled and stable trust region for policy updates in language model training. Developed by the Qwen team, GSPO demonstrates notably superior scalability and training stability compared to token-level methods, and serves as the core RL algorithm underlying the Qwen3 model series.

DGPO \cite{dai2026harder} proposes a difficulty-aware extension of GRPO that dynamically adjusts the optimization process based on the estimated difficulty of each training problem. By incorporating difficulty signals into the advantage computation and combining them with a multi-aspect question reformulation strategy, DGPO steers the policy to allocate greater learning effort toward harder problems, thereby improving the efficiency and effectiveness of RL-based mathematical reasoning training.

\section{Implementation Details}\label{appx:d}

To avoid overfitting caused by adopting a fixed summarization phrase during branching, we maintain a pool of summarization-like token candidates. In the summarization detection phase for NPs, as long as the segment successfully matches any element in the pool, we identify the current position as a natural summarization point. In the summarization injection phase for APs, we randomly select an element from the pool and append it to the truncated prefix. Case-insensitive matching is uniformly applied during character matching. Since our methodology is inherently applicable to arbitrary reasoning paradigms, \textit{we have also incorporated terms such as 'reflection', ``alternatively'', and ``wait'' into the collection}. The full collection of summarization-like words are shown as follows:

\begin{tcolorbox}[
  colback=gray!10,     
  colframe=black,      
  boxrule=1pt,         
  arc=2mm,             
  left=4pt,right=4pt,top=4pt,bottom=4pt, 
] 
``in summary,'', 
``to summarize my progress so far,'', 
``let me summarize what I have so far.'', 
``let me take stock of what I have.'', 
``let me step back and review.'', 
``let me recap what I have done so far.'', 
``putting it all together,'', 
``let me consolidate my work so far.'', 
``to summarize'', 
``let me summarize'', 
``i\'ll summarize'', 
``i will summarize'', 
``summarizing'', 
``in summary'', 
``a quick summary'', 
``brief summary'', 
``quick recap'', 
``let me give a summary'', 
``to recap'', 
``let me recap'', 
``to review'', 
``let me review'', 
``reviewing'', 
``let me re-examine'', 
``re-examining'', 
``let me revisit'', 
``revisiting'', 
``let me re-check'', 
``let me double-check'', 
``double-checking'', 
``let me go back'', 
``going back to'', 
``what i have so far'', 
``what we have so far'', 
``so far i have'', 
``so far, i have'', 
``so far we have'', 
``so far, we have'', 
``up to now'', 
``thus far'', 
``at this point, i have'', 
``at this point, we have'', 
``let me take stock'', 
``taking stock of'', 
``let me consolidate"'', 
``in conclusion'', 
``to conclude'', 
``let me conclude'', 
``concluding'', 
``putting it all together'', 
``combining the above'', 
``combining these results'', 
``combining everything'', 
``taking stock'', 
``to wrap up'', 
``wrapping up'', 
``reflecting on'', 
``let me reflect'', 
``let me think about'', 
``let me reconsider'', 
``reconsidering'', 
``stepping back'', 
``let me step back'', 
``let me pause'', 
``wait, let me'', 
``actually, let me'', 
``let me verify'', 
``verifying'', 
``let me check'', 
``checking my work'', 
``let me confirm'', 
``confirming'', 
``let me try a different approach'', 
``let me try another approach'', 
``let me try again'', 
``let me start over'', 
``let me restart'', 
``let me redo'', 
``let me rethink'', 
``rethinking'', 
``let me approach this differently'', 
``a different way to think about'', 
``alternatively,'', 
``on second thought'', 
``wait,'', 
``wait.'', 
``hmm,'', 
``hmm.'', 
``hold on,'', 
``hold on.'', 
``actually,'', 
``to put it all together'', 
``to put it another way'', 
``to rephrase'', 
``in other words, let me''.
\end{tcolorbox}

Our prompt for integrating periodical summarization is as follows:

\begin{tcolorbox}[
  colback=gray!10,     
  colframe=black,      
  boxrule=1pt,         
  arc=2mm,             
  left=4pt,right=4pt,top=4pt,bottom=4pt, 
] 

 \textit{Please reason step by step. As you work through the problem, periodically summarize your progress and reflect on your approach to make sure you are on the right track. Put your final answer within} \verb|\boxed{}|.
 
\end{tcolorbox}

Following MathForge~\cite{dai2026harder}, our codebase is built upon the Open-R1 framework~\cite{face2025open} and all experiments are conducted on 8 NVIDIA H20 GPUs. For reinforcement learning, responses are generated with a temperature of 1.0 and a maximum completion length of 4096 tokens. During evaluation, we use a temperature of 0.6, a top-$p$ of 0.95, and a maximum generation length of 4096 tokens. The task reward is binary, with a correct final answer yielding 1 and an incorrect answer yielding 0.

For experiments on the MATH dataset, the system prompt instructs the model to reason step by step and enclose the final answer within \verb|\boxed{}|. The maximum prompt length is set to 512 tokens. We generate $G=16$ responses per query and use a training batch size of 32. For ReSum, the rollout budget $B=TJ$ is set to $G$, ensuring a fair comparison of rollout budget with GRPO. For ReSum, the format reward weights for NPs and APs in Eq.~\ref{eq:reward} are set to $0.2$ and $0.05$, respectively. Branching is only applied to the initial rollouts. $\beta$ is set to 0, and the clipping range parameter $\epsilon$ is set to 0.2. We train for 1 epoch with a learning rate of $1{\times}10^{-6}$ and gradient accumulation steps of 1, resulting in 230 training steps. When combining ReSum with GPG, DAPO, or GSPO, we retain the respective baseline's original hyperparameters (e.g., DAPO's asymmetric clipping bounds $\epsilon_{\mathrm{low}}$ and $\epsilon_{\mathrm{high}}$, GSPO's sequence-level importance sampling) and only replace their advantage estimation with our summarization-aware group-relative advantage defined in Eq.~\ref{eq:sumnosum}.

For the multimodal experiments, we extend the maximum prompt length to 2048 tokens and adjust the system prompt to instruct the model to provide the final answer without units within \verb|\boxed{}|. The model is trained for 2 epochs with a learning rate of $1{\times}10^{-6}$, a batch size of 32, and gradient accumulation steps of 1, resulting in 480 training steps. All other settings remain consistent with the text-only experiments.

\begin{table}[t]
\caption{Performance comparison between ReSum and prompt refinement on 6 challenging mathematical reasoning datasets. The best results are in \textbf{bold}. }
\centering
\label{tab:my-table:promptrefinement}
\renewcommand{\arraystretch}{1.1}   
\setlength\tabcolsep{7pt} 
\resizebox{1\columnwidth}{!}{
\begin{tabular}{lccccccc}
\hline
Model  & AIME24 & AIME25 & AMC23 & MATH500 & Minerva & Olympiad & Avg. \\ \hline
\multicolumn{8}{l}{\textit{\textbf{Qwen2.5-Math-1.5B}}} \\
Baseline & 6.87 &	3.65 	&30.94 	&34.95 &	8.55 	&21.93 	& 17.82  \\
Prompt   Refinement & 6.67 & 4.17 & 34.33 & 39.85 & 11.31 & 23.41 & 19.96 \\
\rowcolor{gray!20}
ReSum & \textbf{15.96} & \textbf{6.83}& \textbf{51.33} & \textbf{67.12} & \textbf{24.58} & \textbf{32.59} & \textbf{33.07} \\ \hline
\multicolumn{8}{l}{\textit{\textbf{Qwen2.5-3B}}} \\
Baseline & 2.81 &	0.73& 	22.66 &	48.65 &	13.69 	&19.37 &	17.99   \\
Prompt   Refinement & 2.81 & 1.15 & 22.73 & 47.25 & 14.89 & 17.74 & 17.76 \\
\rowcolor{gray!20}
ReSum & \textbf{7.50 }& \textbf{2.29}& \textbf{39.06} & \textbf{68.40} & \textbf{26.69} & \textbf{28.93} & \textbf{28.81} \\ \hline
\multicolumn{8}{l}{\textit{\textbf{DeepSeek-Math-7B}}} \\ 
Baseline & 0.42  & 0.10  & 13.28 & 31.05 &	9.56 &	9.00 	&10.57  \\
Prompt   Refinement &  0.33 &	0.10 	&13.85 &	30.54 &	9.75 &	10.08 &	10.78   \\
\rowcolor{gray!20}
ReSum & \textbf{1.66}& \textbf{0.96}& \textbf{22.40} & \textbf{43.10} & \textbf{20.22} & \textbf{15.56} & \textbf{17.32} \\ \hline
\end{tabular}}\vspace{-0.6cm}
\end{table}

\begin{table}[t]
\caption{Significance test ($p$-value) of ReSum over DGPO on all backbones of each dataset. }
\centering
\label{tab:my-table:sig}
\renewcommand{\arraystretch}{1.1}   
\setlength\tabcolsep{10pt} 
\resizebox{1\columnwidth}{!}{
\begin{tabular}{lccccccc}
\hline
Model & AIME24 & AIME25 & AMC23 & MATH500 & Minerva & Olympiad & Avg. \\ \hline
Qwen2.5-Math-1.5B & 0.0080 & 0.0121 & 0.0051 & 0.0018 & 0.0020 & 0.0095 & 0.0064 \\
Qwen2.5-3B & 0.0151 & 0.0098 & 0.0353 & 0.0001 & 0.0190 & 0.0076 & 0.0145 \\
DeepSeek-Math-7B & 0.0231 & 0.0049 & 0.0225 & 0.0069 & 0.0028 & 0.0130 & 0.0122 \\
Qwen2.5-Math-7B & 0.0353 & 0.0040 & 0.0077 & 0.0070 & 0.0365 & 0.0008 & 0.0152 \\ \hline
\end{tabular}}\vspace{-0.6cm}
\end{table}

\begin{table}[t]
\caption{Efficiency study of ReSum of each dataset. }
\centering
\label{tab:my-table:eff}
\renewcommand{\arraystretch}{1.1}   
\setlength\tabcolsep{1pt} 
\resizebox{1\columnwidth}{!}{
\begin{tabular}{lccccccccc}
\hline
\multirow{2}{*}{Backbone} & \multicolumn{4}{c}{DGPO} & \multicolumn{5}{c}{ReSum} \\ \cline{2-10} 
 & initial rollout & advantage & train/step & eval/step & initial rollout & branching & advantage & train/step & eval/step \\ \hline
Qwen2.5-Math-1.5B & 21.50 & 0.81 & 22.31 & 12.5 & 10.20 & 14.41 & 0.68 & 25.29 & 12.78 \\ \hline
\end{tabular}}\vspace{-0.3cm}
\end{table}

\section{Further Analysis}\label{appx:e}
\paragraph{Comparisons with naive prompt refinement.} Table~\ref{tab:my-table:promptrefinement} presents the accuracy of the Baseline, Prompt Refinement (i.e., explicitly adding instructions for periodic summarization to the prompt, as shown above), and ReSum across three backbones. We observe that Prompt Refinement yields no substantial change compared to the Baseline, and even underperforms it in some settings. We attribute this to the fact that models without cold-start fine-tuning fail to recognize the importance of summarization when encountering errors or redundant generation; consequently, modifications to the prompt yield negligible improvements. In contrast, although ReSum also forgoes cold-start fine-tuning, it keenly captures the nascent signals of summarization behavior and reinforces this behavior through unique contrastive branching. This ultimately enables the model to emerge with a correct understanding of when and how to summarize.

\paragraph{Significance Test.} We provide the significance test in Table \ref{tab:my-table:sig}. We find that ReSum consistently achieves statistically significant differences over DGPO across all backbones and benchmarks, with all p-values below 0.05. More specifically, for Qwen2.5-Math-1.5B, the p-values range from 0.0018 to 0.0121, with an average of 0.0064; for Qwen2.5-3B, they range from 0.0001 to 0.0353, with an average of 0.0145. Similar trends are observed on DeepSeek-Math-7B and Qwen2.5-Math-7B, whose average p-values are 0.0122 and 0.0152 respectively. Notably, the significance remains stable not only on in-distribution benchmarks such as MATH500, but also on more challenging competition-style datasets such as AIME24, AIME25, and Olympiad. These results suggest that the gains of ReSum are not caused by random fluctuations from repeated sampling, but reflect a robust and reproducible advantage over DGPO. Combined with the consistent improvements in the main results, the significance test further validates that incentivizing periodic self-summarization indeed provides a reliable benefit for long-chain reasoning.

\paragraph{Efficiency.} In Table~\ref{tab:my-table:eff}, we report the average time cost of DGPO and ReSum across all training steps under identical computational budgets. ReSum exhibits an approximate 13\% increase in overall overhead compared to DGPO, which is primarily attributable to the additional costs associated with the identification and execution of branching. Furthermore, we observe that generation volatility during the early training phase is the main driver of this increased cost. As training progresses, the rate of overhead growth steadily declines. This is because the model learns to perform summarization at optimal moments and positions, thereby substantially reducing the overall rollout length.

\paragraph{Case Study.} 
We provide some cases in this part, which shows that ReSum incentivizes the appropriate summarization ability by summarizing   and rethinking generated contents:

\definecolor{sumcolor}{RGB}{255, 200, 0}
\definecolor{errorcolor}{RGB}{255, 100, 100}
\definecolor{correctcolor}{RGB}{100, 200, 100}
\definecolor{codebackground}{RGB}{245, 245, 245}
\definecolor{codeborder}{RGB}{200, 200, 200}
 
\lstset{
    backgroundcolor=\color{codebackground},
    basicstyle=\ttfamily\small,
    frame=single,
    rulecolor=\color{codeborder},
    breaklines=true,
    columns=fullflexible,
}

\begin{tcolorbox}[
    enhanced,
    colback=white, colframe=black!30,
    left=8pt, right=8pt, top=6pt, bottom=6pt,
]
To solve the equation $\sqrt{x+3} + \sqrt{x-2} = 5$, we can follow these steps:

1. Isolate one of the square roots.
2. Square both sides to eliminate the square root.
3. Simplify and isolate the remaining square root.
4. Square both sides again to eliminate the remaining square root.
5. Solve the resulting equation for $x$.
6. Verify the solution by substituting back into the original equation.

Let's implement this step-by-step in Python using SymPy:
\begin{lstlisting}[language=Python]
import sympy as sp

x = sp.symbols('x')
equation = sp.sqrt(x + 3) + sp.sqrt(x - 2) - 5
solution = sp.solve(equation, x)
valid_solutions = [sol for sol in solution if equation.subs(x, sol) == 0]
print(valid_solutions)
\end{lstlisting}
\begin{lstlisting}
[7]
\end{lstlisting}

The solution to the equation $\sqrt{x+3} + \sqrt{x-2} = 5$ is $x = 7$. To verify, we can substitute $x = 7$ back into the original equation:
\[
\sqrt{7+3} + \sqrt{7-2} = \sqrt{10} + \sqrt{5}
\]
We need to check if $\sqrt{10} + \sqrt{5} = 5$. Let's compute this explicitly:
\[
\sqrt{10} \approx 3.162, \quad \sqrt{5} \approx 2.236, \quad \sqrt{10} + \sqrt{5} \approx 5.398 \neq 5
\]

\colorbox{sumcolor}{\textbf{To summarize}}, it seems there was an error in the verification. Let's recheck the solution by solving the equation again carefully.

We'll start by isolating one of the square roots and then squaring both sides:
\[
\sqrt{x+3} = 5 - \sqrt{x-2}
\]
\[
x + 3 = (5 - \sqrt{x-2})^2 = 25 - 10\sqrt{x-2} + (x-2)
\]
\[
3 = 23 - 10\sqrt{x-2} \implies 10\sqrt{x-2} = 20 \implies \sqrt{x-2} = 2
\]
\[
x - 2 = 4 \implies x = 6
\]

Let's verify $x = 6$ by substituting back into the original equation:
\[
\sqrt{6+3} + \sqrt{6-2} = \sqrt{9} + \sqrt{4} = 3 + 2 = 5 \checkmark
\]

This verifies that $x = 6$ is the correct solution. The previous output of 7 was incorrect due to a verification error.
\begin{lstlisting}[language=Python]
valid_solutions = [sol for sol in solution if equation.subs(x, sol) == 0]
print(valid_solutions)
\end{lstlisting}
\begin{lstlisting}
[6]
\end{lstlisting}

The final answer is:
\[
\boxed{6}
\]
\end{tcolorbox}

\begin{tcolorbox}[
    enhanced,
    colback=white, colframe=black!30,
    left=8pt, right=8pt, top=6pt, bottom=6pt,
]

\textbf{Problem:} How many 4-digit numbers have digits that sum to 10, where each digit is between 1 and 4 inclusive?

\tcblower

\textbf{Model Reasoning:}

\medskip
To solve this problem, we need to find the number of solutions to
\[
    x_1 + x_2 + x_3 + x_4 = 10, \quad 1 \leq x_i \leq 4.
\]

1. Variable substitution.
Let $y_i = x_i - 1$. Then $0 \leq y_i \leq 3$ and the equation becomes
\[
    y_1 + y_2 + y_3 + y_4 = 6.
\]

2. Unconstrained count. 
By stars and bars, the total number of non-negative integer solutions is
\[
    \binom{6 + 4 - 1}{4 - 1} = \binom{9}{3} = 84.
\]

3. Subtract violating cases.
If $y_1 > 3$, let $z_1 = y_1 - 4 \geq 0$. The equation becomes $z_1 + y_2 + y_3 + y_4 = 2$, giving $\binom{5}{3} = 10$ solutions. By symmetry, there are $4 \times 10 = 40$ violating cases.

4. Check double-counted cases.
If $y_1 > 3$ and $y_2 > 3$ simultaneously, the equation becomes $z_1 + z_2 + y_3 + y_4 = -2$, which has \emph{no solutions}. No correction needed.

\medskip
\parbox{\dimexpr\linewidth-2\fboxsep}{%
\colorbox{sumcolor}{\textbf{To summarize}}
}

\medskip
Conclusion.
Putting it all together, the number of valid solutions is
\[
    \binom{9}{3} - 4 \cdot \binom{5}{3} = 84 - 40 = \boxed{44}.
\]

\medskip
Code verification:
\begin{verbatim}
import math
binom_9_3 = math.comb(9, 3)   # 84
binom_5_3 = math.comb(5, 3)   # 10
valid_solutions = binom_9_3 - 4 * binom_5_3
print(valid_solutions)         # Output: 44
\end{verbatim}

\medskip
The final answer is: $\boxed{44}$

\end{tcolorbox}

\begin{tcolorbox}[
    enhanced,
    colback=white, colframe=black!30,
    left=8pt, right=8pt, top=6pt, bottom=6pt,
]
To solve the equation $\lfloor x \rfloor + \lfloor 2x \rfloor = 5$, let's break down the problem step-by-step.

1. Let $\lfloor x \rfloor = n$, where $n$ is an integer. This means that $n \leq x < n+1$.
2. Then, $\lfloor 2x \rfloor$ depends on the value of $2x$. Since $n \leq x < n+1$, we have $2n \leq 2x < 2n+2$. Therefore, $\lfloor 2x \rfloor$ can be either $2n$ or $2n+1$.
3. We need to satisfy the equation $n + \lfloor 2x \rfloor = 5$.
4. This gives us two cases to consider:
   \begin{itemize}
       \item Case 1: $\lfloor 2x \rfloor = 2n$. Then, $n + 2n = 5$, so $3n = 5$, which gives $n = \frac{5}{3}$. Since $n$ must be an integer, this case is not possible.
       \item Case 2: $\lfloor 2x \rfloor = 2n+1$. Then, $n + (2n + 1) = 5$, so $3n + 1 = 5$, which gives $3n = 4$, and thus $n = \frac{4}{3}$. Since $n$ must be an integer, this case is also not possible for the same reason.
   \end{itemize}
\colorbox{sumcolor}{\textbf{To summarize}} both possible cases lead to non-integer values of $n$. Since $n=\lfloor x \rfloor$ must be an integer, neither case can occur. Therefore, the equation has no real solution.

Let's verify this conclusion by checking a dense grid of real values.
\begin{lstlisting}[language=Python]
import numpy as np

def check_solution(x):
    return np.floor(x) + np.floor(2 * x) == 5

test_points = np.linspace(-10, 10, 10000)
solutions = [x for x in test_points if check_solution(x)]
print(solutions)
\end{lstlisting}
\begin{lstlisting}
[]
\end{lstlisting}

The Python check finds no sampled real values satisfying the equation, which is consistent with the case analysis.

The final answer is:
\[
\boxed{\text{No real solution}}
\]
\end{tcolorbox}

\section{Algorithm}\label{appx:f}
The algorithm of ReSum is shown in algorithm~\ref{alg:resum}.

\DontPrintSemicolon
\SetKwInOut{Input}{Input}
\SetKwProg{Fn}{Function}{}{end}
\SetKwComment{tcp}{// }{}
\SetAlCapSkip{0.5em}
\SetAlgoNlRelativeSize{-1}
\SetAlFnt{\small}
\SetKw{KwTo}{to}

\begin{algorithm}[t]
\caption{ReSum Training Pipeline}
\label{alg:resum}
\Input{policy $\pi_\theta$, reference policy $\pi_{\mathrm{ref}}$, training set $\mathcal{D}$,
       number of trees $T$, generation loops $J$, rollout budget $B=TJ$ (matched to GRPO group size $G$),
       summarization phrase pool $\mathcal{P}$, clipping $\epsilon$, KL coefficient $\beta$,
       advantage stabilizer $\xi>0$}
\tcp{$J$: generation loops; $j{=}1$ samples initial rollouts; $B{=}TJ$ total rollouts.}

\For{each training step}{
  Sample query $q \sim \mathcal{D}$ and freeze $\pi_{\mathrm{old}} \leftarrow \pi_\theta$\;

  \For{$t = 1$ \KwTo $T$}{
    $\mathcal{T}_t \leftarrow \emptyset$\;
  }

  \tcp{--- Rollout generation: $j=1,\ldots,J$; one rollout per tree per loop ---}
  \For{$j = 1$ \KwTo $J$}{
    \For{$t = 1$ \KwTo $T$}{
      \uIf{$j = 1$}{
        $\tau_{t,1} \sim \pi_{\mathrm{old}}(\cdot \mid q)$, $\mathrm{kind}(\tau_{t,1}) \leftarrow \texttt{root}$\;
        $\mathcal{N}_t \leftarrow$ ordered NP positions on $\tau_{t,1}$ detected by $\mathcal{P}$\;
        $\mathcal{A}_t \leftarrow$ non-summary token positions on $\tau_{t,1}$
        (excluding $\mathcal{N}_t$ and other phrase matches)\;
        store $\tau_{t,0} \leftarrow \tau_{t,1}$ as the branching anchor\;
      }{
        \tcp{branch only from the initial rollout $\tau_{t,0}$}
        \eIf{$\mathcal{N}_t \neq \emptyset$}{
          pop the next NP position $p$ from $\mathcal{N}_t$\;
          $\mathcal{F} \leftarrow$ prefix of $\tau_{t,0}$ up to $p$, with the matched phrase masked\;
          $\tau_{t,j} \sim \pi_{\mathrm{old}}(\cdot \mid q, \mathcal{F})$, $\mathrm{kind}(\tau_{t,j}) \leftarrow \texttt{np}$\;
        }{
          sample AP position $p$ uniformly from $\mathcal{A}_t$\;
          $\mathcal{F} \leftarrow$ prefix of $\tau_{t,0}$ up to $p$, appended with $s \sim \mathcal{P}$\;
          $\tau_{t,j} \sim \pi_{\mathrm{old}}(\cdot \mid q, \mathcal{F})$, $\mathrm{kind}(\tau_{t,j}) \leftarrow \texttt{ap}$\;
        }
      }
      $\mathcal{T}_t \leftarrow \mathcal{T}_t \cup \{\tau_{t,j}\}$\;
    }
  } 

  \tcp{--- Rewards (Eq.~\eqref{eq:reward}) ---}
  \For{$\tau \in \bigcup_{t=1}^{T} \mathcal{T}_t$}{
    $R_A(\tau) \leftarrow \mathbf{1}[\mathrm{Correct}(\tau)]$, $n \leftarrow \#\{\text{phrase matches from }\mathcal{P}\text{ in }\tau\}$\;
    \uIf{$\mathrm{kind}(\tau) = \texttt{ap}$}{
      $N_{\mathrm{NP}}(\tau) \leftarrow n$;\quad $N_{\mathrm{AP}}(\tau) \leftarrow 1$\;
    }{
      $N_{\mathrm{NP}}(\tau) \leftarrow n$;\quad $N_{\mathrm{AP}}(\tau) \leftarrow 0$\;
    }
    \eIf{$N_{\mathrm{NP}}(\tau) + N_{\mathrm{AP}}(\tau) > 0$}{
      $R_F(\tau) \leftarrow
      \dfrac{0.2\,N_{\mathrm{NP}}(\tau) + 0.05\,N_{\mathrm{AP}}(\tau)}
            {N_{\mathrm{NP}}(\tau) + N_{\mathrm{AP}}(\tau)}$\;
    }{
      $R_F(\tau) \leftarrow 0$\;
    }
    $R(\tau) \leftarrow R_A(\tau) + R_F(\tau)$; store $n(\tau) \leftarrow n$\;
  }

  \tcp{--- Summarization-aware advantages (Eq.~\eqref{eq:sumnosum}) ---}
  $G_{\mathrm{sum}} \leftarrow
    \{\tau : n(\tau) > 0 \;\vee\; \mathrm{kind}(\tau) = \texttt{ap}\}$; $G_{\mathrm{nosum}} \leftarrow
    \{\tau \in \textstyle\bigcup_t \mathcal{T}_t : \tau \notin G_{\mathrm{sum}}\}$\;

  \If{$G_{\mathrm{sum}} \neq \emptyset$}{
    $\mu_{\mathrm{sum}} \leftarrow \mathrm{mean}_{\tau \in G_{\mathrm{sum}}} R(\tau)$; $\sigma_{\mathrm{sum}} \leftarrow \mathrm{std}_{\tau \in G_{\mathrm{sum}}} R(\tau)$\;
  }
  \If{$G_{\mathrm{nosum}} \neq \emptyset$}{
    $\mu_{\mathrm{nosum}} \leftarrow \mathrm{mean}_{\tau \in G_{\mathrm{nosum}}} R(\tau)$; $\sigma_{\mathrm{nosum}} \leftarrow \mathrm{std}_{\tau \in G_{\mathrm{nosum}}} R(\tau)$\;
  }

  \For{$\tau \in \bigcup_{t=1}^{T} \mathcal{T}_t$}{
    $\hat{A}^{\mathrm{sum}}(\tau) \leftarrow
    \begin{cases}
      \dfrac{R(\tau) - \mu_{\mathrm{sum}}}{\sigma_{\mathrm{sum}} + \xi},
        & G_{\mathrm{sum}} \neq \emptyset \\
      0, & \text{otherwise}
    \end{cases}$\;
    $\hat{A}^{\mathrm{nosum}}(\tau) \leftarrow
    \begin{cases}
      \dfrac{R(\tau) - \mu_{\mathrm{nosum}}}{\sigma_{\mathrm{nosum}} + \xi},
        & G_{\mathrm{nosum}} \neq \emptyset \\
      0, & \text{otherwise}
    \end{cases}$\;
    $\hat{A}(\tau) \leftarrow \hat{A}^{\mathrm{sum}}(\tau) + \hat{A}^{\mathrm{nosum}}(\tau)$\;
  }

  \tcp{--- SGPO policy update (Eq.~\eqref{eq:obj}) ---}
  \For{each $\tau_{t,e} \in \bigcup_{t=1}^{T} \mathcal{T}_t$}{
    \For{$i = 1$ \KwTo $|\tau_{t,e}|$}{
      $\rho_{t,e,i}(\theta) \leftarrow
        \dfrac{\pi_\theta(\tau_{t,e,i} \mid \tau_{t,e,<i}, q)}
              {\pi_{\mathrm{old}}(\tau_{t,e,i} \mid \tau_{t,e,<i}, q)}$\;
    }
  }
  Update $\theta$ by maximizing $J_{\mathrm{ReSum}}(\theta)$,
  where each token in $\tau_{t,e}$ uses rollout-level $\hat{A}(\tau_{t,e})$\;
}
\end{algorithm}
  
\section{Impact Statement}\label{appx:g}

ReSum improves reasoning efficiency by internalizing self-summarization, reducing unnecessary computation and output length, which can lower energy costs for large-scale deployment. By keeping reasoning contexts concise and faithful, it may enhance the reliability of AI systems in education and science. We foresee no direct negative societal consequences unique to this method, though general concerns about language model misuse persist. Overall, the societal benefits of more efficient and trustworthy reasoning outweigh potential risks.

\section{Limitations}\label{appx:h}
The tree-structured rollout of ReSum  adds implementation complexity in managing branching points and prefix processing, though the total rollout budget per query remains the same as the baselines. The detection of NPs relies on simple keyword matching, which could miss implicit or domain-specific summarization patterns. Our current evaluation focuses primarily on mathematical reasoning and one multimodal dataset; the effectiveness of ReSum on other tasks, such as code generation or open-ended language tasks, is not yet studied. Hyperparameters \(T\) and \(J\) require manual tuning across model scales and budgets. Overall, our design and validation of ReSum strike an optimal balance between efficiency and effectiveness.

\section{Use of LLMs}\label{appx:i}
LLMs are only for polishing the writing of this paper.

\end{document}